\documentclass[10pt,twocolumn,letterpaper]{article}

\usepackage{iccv}
\usepackage{times}
\usepackage{epsfig}
\usepackage{graphicx}
\usepackage{amsmath}
\usepackage{amssymb}

\usepackage{booktabs}
\usepackage{comment}
\usepackage{subfigure}
\usepackage{multirow}

\usepackage[pagebackref=true,breaklinks=true,letterpaper=true,colorlinks,bookmarks=false]{hyperref}

\iccvfinalcopy 

\def\iccvPaperID{10057} 

\ificcvfinal\fi

\begin{document}

\title{BioFors: A Large Biomedical Image Forensics Dataset}

\author{Ekraam Sabir, Soumyaroop Nandi, Wael AbdAlmageed, Prem Natarajan\\
USC Information Sciences Institute, Marina del Rey, CA, USA\\
{\tt\small \{esabir, soumyarn, wamageed, pnataraj\}@isi.edu}
}

\maketitle

\begin{abstract}
   Research in media forensics has gained traction to combat the spread of misinformation. However, most of this research has been directed towards content generated on social media. Biomedical image forensics is a related problem, where manipulation or misuse of images reported in biomedical research documents is of serious concern. The problem has failed to gain momentum beyond an academic discussion due to an absence of benchmark datasets and standardized tasks. In this paper we present BioFors -- the first dataset for benchmarking common biomedical image manipulations. BioFors comprises 47,805 images extracted from 1,031 open-source research papers. Images in BioFors are divided into four categories -- Microscopy, Blot/Gel, FACS and Macroscopy. We also propose three tasks for forensic analysis -- external duplication detection, internal duplication detection and cut/sharp-transition detection. We benchmark BioFors on all tasks with suitable state-of-the-art algorithms. Our results and analysis show that existing algorithms developed on common computer vision datasets are not robust when applied to biomedical images, validating that more research is required to address the unique challenges of biomedical image forensics.
\end{abstract}
\section{Introduction}


\begin{figure}
    \centering
    \includegraphics[width=\linewidth]{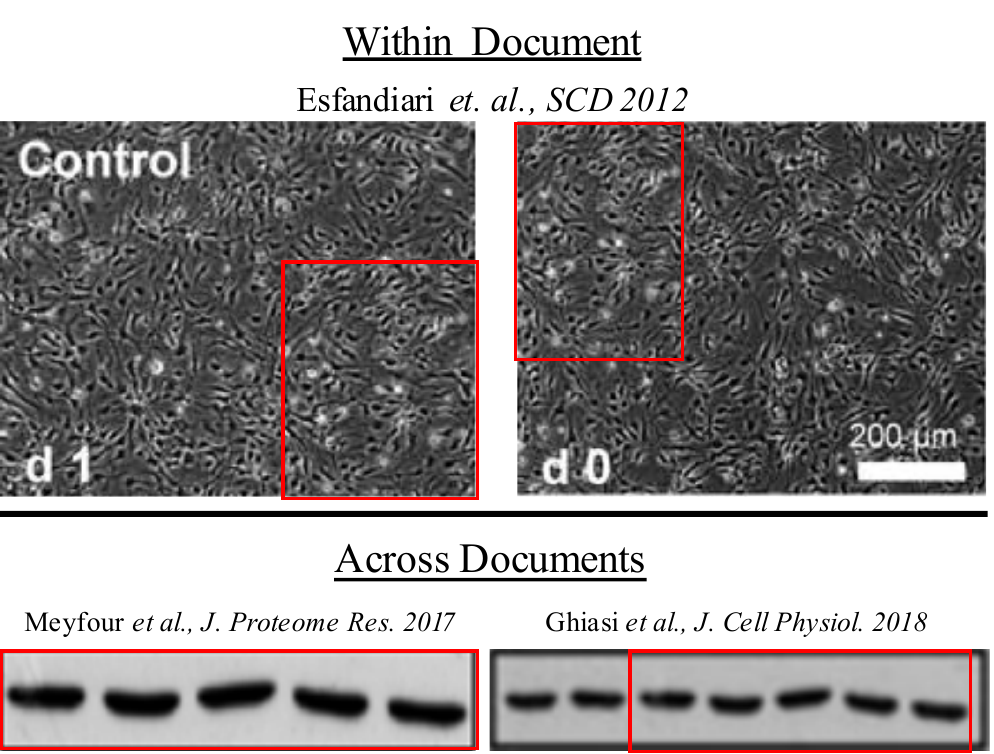}
    \caption{Real world examples of suspicious duplications in biomedical images. Top and bottom rows show duplications between images in the same and different documents respectively.}
    \label{teaser}
\end{figure}

Multimedia forensic research has branched off into several sub-domains to tackle various forms of misinformation and manipulation. Popular forensic research problems include detection of digital forgeries such as deepfakes \cite{masi2020two,sabir2019recurrent}, copy-move and splicing manipulations \cite{Wu_2018_ECCV,Wu_2019_CVPR,wu2017deep} or semantic forgeries \cite{sabir2018deep,jaiswal2019aird}. These  forensic-research areas essentially deal with social media content. A related but distinct research domain is biomedical image forensics; i.e. detection of research misconduct in biomedical publications \cite{bik2016prevalence,christopher2018systematic,bik2018analysis}. Research misconduct can appear in several forms such as plagiarism, fabrication and falsification. Scientific misconduct has consequences beyond ethics and leads to retractions \cite{bik2018analysis} and by one estimate $\textdollar392,582$ of financial loss for each retracted article \cite{stern2014financial}. The general scope of scientific misconduct and unethical behavior is broad. In this paper we focus on detection of manipulation or inappropriate duplication of scientific images in biomedical literature.

Duplication and tampering of protein, cell, tissue and other experimental images has become a nuisance in the biomedical sciences community. As the description suggests, duplication involves reusing part of images generated by one experiment to misrepresent results for unrelated experiments. Tampering of images involves pixel- or patch-level forgery to hide unfavorable aspects of the image or to produce favorable results. Biomedical image forgeries can be more difficult for a human to detect than manipulated images on social media due to the presence of arbitrary and confusing patterns and lack of real-world semantic context. Detecting forgeries is further complicated by manipulations involving images across different documents. Figure \ref{teaser} shows reported examples\footnote{\href{https://scienceintegritydigest.com/2020/11/11/46-papers-from-a-royan-institute-professor/}{\textcolor{red}{https://scienceintegritydigest.com/2020/11/11/46-papers-from-a-royan-institute-professor/}}} of inappropriate duplications in different publications. The difficulty of noticing such manipulations coupled with a high paper-per-reviewer ratio often leads to these manipulations going unnoticed during the review process. It may come under scrutiny later leading to possible retractions \cite{bik2018analysis}. While the problem has received the attention of the biomedical community, to the best of our knowledge there is no publicly available biomedical image forensics dataset, detection software or standardized task for benchmarking. We address these issues by releasing the first biomedical image forensics dataset (BioFors) and proposing benchmarking tasks. 

 The objective of our work is to advance biomedical forensic research to identify suspicious images with high confidence. We hope that BioFors will promote the development of algorithms and software which can help reviewers identify manipulated images in research documents. The final decision regarding malicious, mistaken or justified intent behind a suspicious image is to be left to the forensic analyst. This is important due to cases of duplication/tampering that are justified with citation, explanation, harmlessness or naive mistake as detailed in \cite{bik2016prevalence}. BioFors comprises 47,805 manually cropped images belonging to four major categories --- (1) Microscopy, (2) Blot/Gel, (3) Macroscopy and (4) Flow-cytometry or Fluoroscence-activated cell sorting (FACS). It covers popular biomedical image manipulations with three forgery detection tasks. The dataset and its collection along with the forgery detection tasks are detailed in Section \ref{dataset}. 

The contributions of our work are:
\begin{itemize}
    \item A large scale biomedical image forensics dataset with real-world forgeries
    \item A computation friendly taxonomy of forgery detection tasks that can be matched with standard computer vision tasks for benchmarking and evaluation
    \item Extensive analysis explaining the challenges of biomedical forensics and the loss in performance of standard computer vision models when applied to biomedical images
\end{itemize}

\section{Related Work}\label{related}

\subsection{Computer Vision in Biomedical Domain}

Machine learning and computer vision have made significant contributions to the biomedical domain involving problems such as image segmentation \cite{Kulikov_2020_CVPR,Lee_2020_CVPR,Wang_2020_CVPR}, disease diagnostics \cite{Perez_2019_CVPR_Workshops}, super-resolution \cite{Peng_2020_CVPR} and biomedical image denoising \cite{zhang2019poisson}. While native computer vision algorithms have existed for these problems, ensuring robustness on biomedical data has always been a challenge. This is partly due to domain shift and also due to the  difficulty of training data-intensive deep learning models on biomedical datasets which are usually small. 

\subsection{Natural-Image Forensics}\label{natural_forensics}

Image forensics is a widely studied problem in computer vision with standard datasets and benchmarking \cite{verdoliva2020media}. Common forensics problems include deepfake detection \cite{masi2020two,sabir2019recurrent}, splicing \cite{wu2017deep,cozzolino2015splicebuster}, copy-move forgery detection (CMFD) \cite{Wu_2018_ECCV,cozzolino2015efficient,ryu2010detection}, enhancement and removal detection \cite{Wu_2019_CVPR,Zhou_2018_CVPR}. While some forms of manipulation such as image enhancement may be harmless, others have malicious intent. Recently, deepfakes -- a class of forgeries where a person's identity or facial experession is manipulated, has gained notoriety. Other malicious forms of forgeries are copy-move and splicing which involve pasting an image patch from within the same image and from a donor image respectively. For the manipulations mentioned, forgery detection methods have been developed to flag suspicious content with reasonable success. A critical step for the development of these algorithms has been the curation and release of datasets that facilitated benchmarking. As an example, FF++ \cite{rossler2019faceforensics++}, DeeperForensics \cite{jiang2020deeperforensics} and Celeb-DF \cite{Li_2020_CVPR} helped develop methods for deepfake detection. Similarly, CASIA \cite{dong2013casia}, NIST16 \cite{NimbleCh33:online}, COLUMBIA \cite{ng2009columbia} and COVERAGE \cite{wen2016coverage} helped advance detection methods for a combination of forgeries such as copy-move, splicing and removal. 

\subsection{Biomedical-Image Forensics}

Misrepresentation of scientific research is a broad problem \cite{boutron2018misrepresentation} out of which image manipulation or duplication of biomedical images has been recognized as a serious problem by journals and the community in general \cite{christopher2018systematic,bik2016prevalence,bik2018analysis}. Bik \textit{et al.} \cite{bik2016prevalence} analyzed over 20,000 papers and found 3.8\% of these to contain at least one manipulation. In continuing research \cite{bik2018analysis}, the authors were able to bring 46 corrections or retractions. However, most of this effort was performed manually which is unlikely to scale given the high  volume of publications. Models and frameworks have been proposed for automated detection of biomedical image manipulations \cite{cardenuto2019scientific,bucci2018automatic,acuna2018bioscience,xiang2020scientific,koppers2017towards}. Koppers \textit{et al.} \cite{koppers2017towards} developed a duplication screening tool evaluated on three images. Bucci \textit{et al.} \cite{bucci2018automatic} engineered a CMFD framework from open-source tools to evaluate 1,546 documents and found 8.6\% of it to contain manipulations. Acuna \textit{et al.} \cite{acuna2018bioscience} used SIFT \cite{lowe2004distinctive} image-matching to find potential duplication candidates in 760k documents, followed by human review. In the absence of a robust evaluation, it is unknown how many documents with forgeries went unnoticed in \cite{bucci2018automatic,acuna2018bioscience}. Cardenuto \textit{et al.} \cite{cardenuto2019scientific} curated a dataset of 100 images to evaluate an end-to-end framework for CMFD task. Xiang \textit{et al.} \cite{xiang2020scientific} test a heterogenous feature extraction model to detect artificially created manipulations in a dataset of 357 microscopy and 487 western blot images. It is unclear how the images were collected in \cite{cardenuto2019scientific,xiang2020scientific}. In summary, none of the proposed datasets unify the community around biomedical image forensics with standard benchmarking.

\section{BioFors Benchmark}\label{dataset}

As discussed in Section \ref{related}, a  dataset with standardized benchmarking is essential to advance the field of biomedical image forensics. Additionally, we want BioFors to have image level granularity in order to facilitate image and pixel level evaluation. Furthermore, it is desirable to use images with real-world manipulations. To this end, we used open-source or retracted research documents to curate BioFors. BioFors is a reasonably large dataset at the intersection of biomedical and image-forensics domain, with 46,064 pristine and 1,741 manipulated images, when compared to biomedical image datasets including  FMD \cite{zhang2019poisson} (12,000 images before augmentation) and CVPPP \cite{scharr2014annotated} (284 images) and also compared to image forensics datasets, including Columbia \cite{ng2009columbia} (180 tampered images), COVERAGE \cite{wen2016coverage} (100 tampered images), CASIA \cite{dong2013casia} (5,123 tampered images) and MFC \cite{guan2019mfc} (100k tampered images). Section \ref{collection} details the image collection procedure. Image diversity and categorization is described in Section \ref{description}. Proposed manipulation detection tasks are described in Section \ref{manipulation_tasks}. A discussion on ethics is included in supplementary material.

\subsection{Image Collection Procedure}\label{collection}

Most research publications do not exhibit forgery, therefore collecting documents with manipulations is a difficult task. We received a set of documents from Bik \textit{et al.} \cite{bik2016prevalence} along with raw annotations of suspicious scientific images which will be discussed in Section \ref{manipulation_tasks}. Of the list of documents from different journals provided to us, we selected documents from PLOS ONE open-source journal comprising 1031 biomedical research documents published between January 2013 and August 2014. 

The collected documents were in Portable Document Format (PDF), however direct extraction of biomedical images from PDF documents is not possible with available software. Furthermore, figures in biomedical documents are compound figures \cite{shi2019layout,tsutsui2017data} i.e. a figure comprises biomedical images, charts, tables and other artifacts. Sadly, state-of-the-art biomedical figure decomposition models \cite{shi2019layout,tsutsui2017data} have imperfect and overlapping crop boundaries. We overcome these challenges in two steps: 1) automated extraction of figures from documents and 2) manual cropping of images from figures. For automated figure extraction we used deepfigures \cite{siegel2018extracting}. We experimented with other open source figure extractors, but deepfigures had significantly better crop boundaries and worked well on all the documents. We obtained 6,543 figure images out of which 5,035 figures had biomedical images. For the cropping step, in order to minimize human error in manual crop boundaries we performed cropping in two stages. We cropped sub-figures with a loose bounding box, followed by a tight crop around images of interest. We filtered out synthetic/computer generated images such as tables, bar plots, histograms, graphs, flowcharts and diagrams. Verification of numerical results in synthetic images is beyond the scope of this paper. The image collection process resulted in 47,805 images. We created the train/test split such that a document and its images belong to the test set if it has at least one manipulation. Table \ref{dataset_breakdown} gives an overview of the dataset. For more statistics on BioFors please refer to the supplementary material.

\begin{table}[h]
    \small
    \centering
    \begin{tabular}{lrrr}
         \toprule
         \textbf{Modality} &  \textbf{Train} & \textbf{Test} & \textbf{Total} \\
         \cmidrule(r){1-1}\cmidrule(l){2-4}
         Documents & 696 & 335 & 1,031 \\
         Figures & 3,377 & 1,658 & 5,035 \\
         All Images & 30,536 & 17,269 & 47,805 \\
         \cmidrule(r){1-1}\cmidrule(l){2-4}
         Microscopy Images & 10,458 & 7,652 & 18,110 \\
         Blot/Gel Images & 19,105 & 8,335 & 27,440 \\
         Macroscopy Images & 555 & 639 & 1,194 \\
         FACS Images & 418 & 643 & 1,061 \\
         \bottomrule
    \end{tabular}
    \caption{Top rows give a high level view of BioFors. Bottom rows provide statistics by image category. Training set comprises pristine images and documents.}
    \label{dataset_breakdown}
\end{table}

\subsection{Dataset Description}\label{description}
\begin{figure*}
    \centering
    \includegraphics[width=\textwidth]{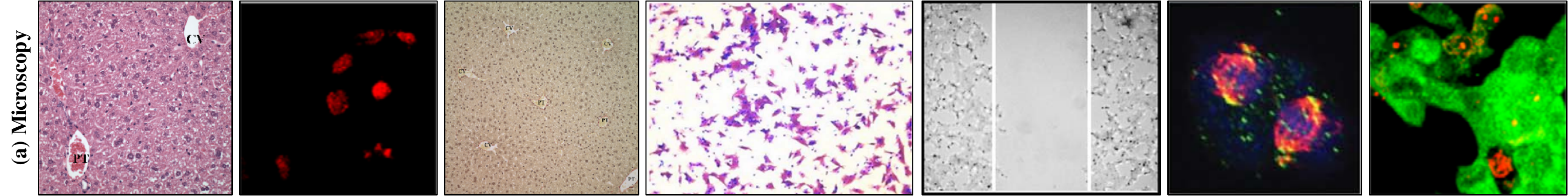}
    \includegraphics[width=\textwidth]{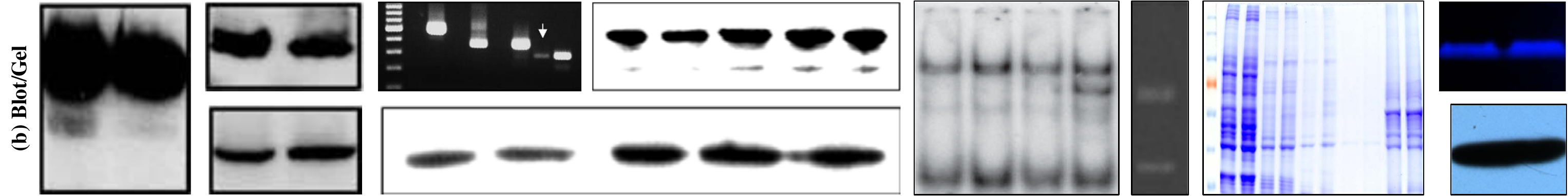}
    \includegraphics[width=\textwidth]{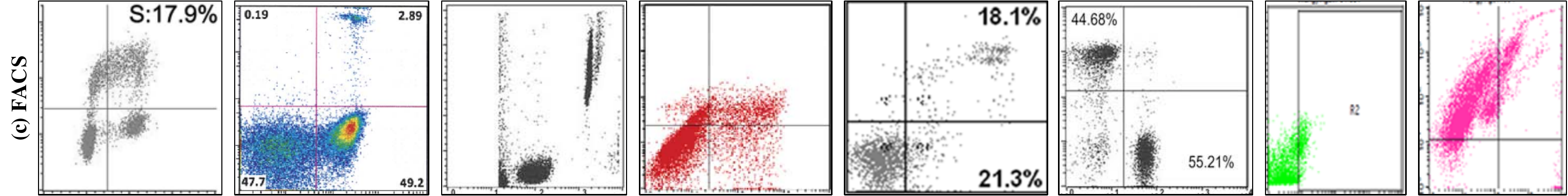}
    \includegraphics[width=\textwidth]{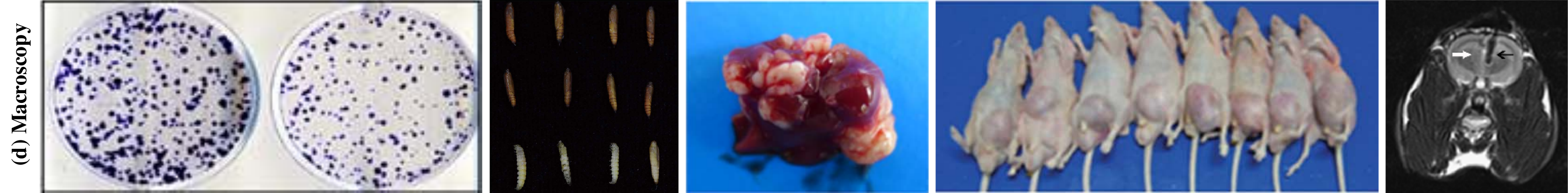}
    \caption{Rows of image samples representative of the following image classes: (a) Microscopy (b) Blot/Gel (c) FACS and (d) Macroscopy.}
    \label{image_samples}
\end{figure*}

We classify the images from the previous collection step into four categories --- (1) Microscopy (2) Blots/Gels (3) Flow-cytometry or Fluoroscence-activated cell sorting (FACS) and (4) Macroscopy. This taxonomy is made considering both the semantics and visual similarity of different image classes. Semantically, microscopy includes images from experiments that are captured using a microscope. They include images of tissues and cells. Variations in microscopy images can result from factors pertaining to origin (e.g. human, animal, organ) or fluorescent chemical staining of cells and tissues. This produces images of diverse colors and structures. Western, northern and southern blots and gels are used for analysis of proteins, RNA and DNA respectively. The images look similar and the specific protein or blot types are visually indistinguishable. FACS images look similar to synthetic scatter plots. However, the pattern is generated by a physical experiment which represents the scattering of cells or particles. Finally, Macroscopy includes experimental images that are visible to the naked eye and do not fall into any of the first three categories. Macroscopy is the most diverse image class with images including rat specimens, tissues, ultrasound, leaves, etc. Table \ref{dataset_breakdown} shows the composition of BioFors by image class. Figure \ref{image_samples} shows inter and intra-class diversity of each class. The image categorization discussed here is easily learnable by popular image classification models as shown in Table \ref{image_classification}. 

\begin{table}[]
    \centering
    \begin{tabular}{lcc}
    \toprule
        \textbf{Model} & \textbf{Train} & \textbf{Test} \\
         \cmidrule(r){1-1}\cmidrule(l){2-3}
         VGG16 \cite{simonyan2014very} & \textbf{99.79\%} & 97.11\% \\
         DenseNet \cite{huang2017densely} & 99.25\% & \textbf{97.67\%} \\
         ResNet \cite{he2016deep} & 98.93\% & 97.47\% \\
         \bottomrule
    \end{tabular}
    \caption{Accuracy of classifying BioFors images using popular image classification models is reliable.}
    \label{image_classification}
\end{table}

\subsection{Manipulation Detection Tasks in BioFors}\label{manipulation_tasks}

\begin{table}[]
\small
    \centering
    \begin{tabular}{lrrr}
    \toprule
          \textbf{Modality} & \textbf{EDD} & \textbf{IDD} & \textbf{CSTD} \\
         \cmidrule(r){1-1}\cmidrule(l){2-4}
         Documents & 308 & 54 & 61 \\
         Pristine Images & 14,675 & 2,307 & 1,534 \\
         Manipulated Images & 1,547 & 102 & 181 \\
         All Images & 16,222 & 2,409 & 1,715 \\
         \bottomrule
    \end{tabular}
    \caption{Distribution of pristine and tampered images in the test set by manipulation task.}
    \label{manipulation_dist}
\end{table}

The raw annotations provided by Bik \textit{et al.} \cite{bik2016prevalence} contain freehand annotations of manipulated regions and notes explaining why the authors of \cite{bik2016prevalence} consider them manipulated. However, the annotation format was not directly useful for ground truth computation. We inspected all suspicious images and manually created binary ground truth masks for all manipulations. This process resulted in 297 documents containing at least one manipulation. We also checked the remaining documents for potentially overlooked manipulations and found another 38 documents with at least one manipulation. Document level cohen's kappa ($\kappa$) inter-rater agreement between biomedical experts (raw annotations) and computer-vision experts (final annotation) is 0.91.

Unlike natural-image forensic datasets \cite{ng2009columbia,wen2016coverage,NimbleCh33:online,dong2013casia} that include synthetic manipulations, BioFors has real-world suspicious images where the forgeries are diverse and the image creators do not share the origin of images or manipulation. Therefore, we do not have the ability to create a one-to-one mapping of biomedical image manipulation detection tasks to the forgeries described in Section \ref{natural_forensics}. Consequently, we propose three manipulation detection tasks in BioFors --- (1) external duplication detection, (2) internal duplication detection and (3) cut/sharp-transition detection. These tasks comprehensively cover the manipulations presented in \cite{bik2016prevalence,christopher2018systematic}. Table \ref{manipulation_dist} shows the distribution of documents and images in the test set across tasks. We describe the tasks and their annotation ahead.

\begin{figure}
    \centering
    \includegraphics[width=\linewidth]{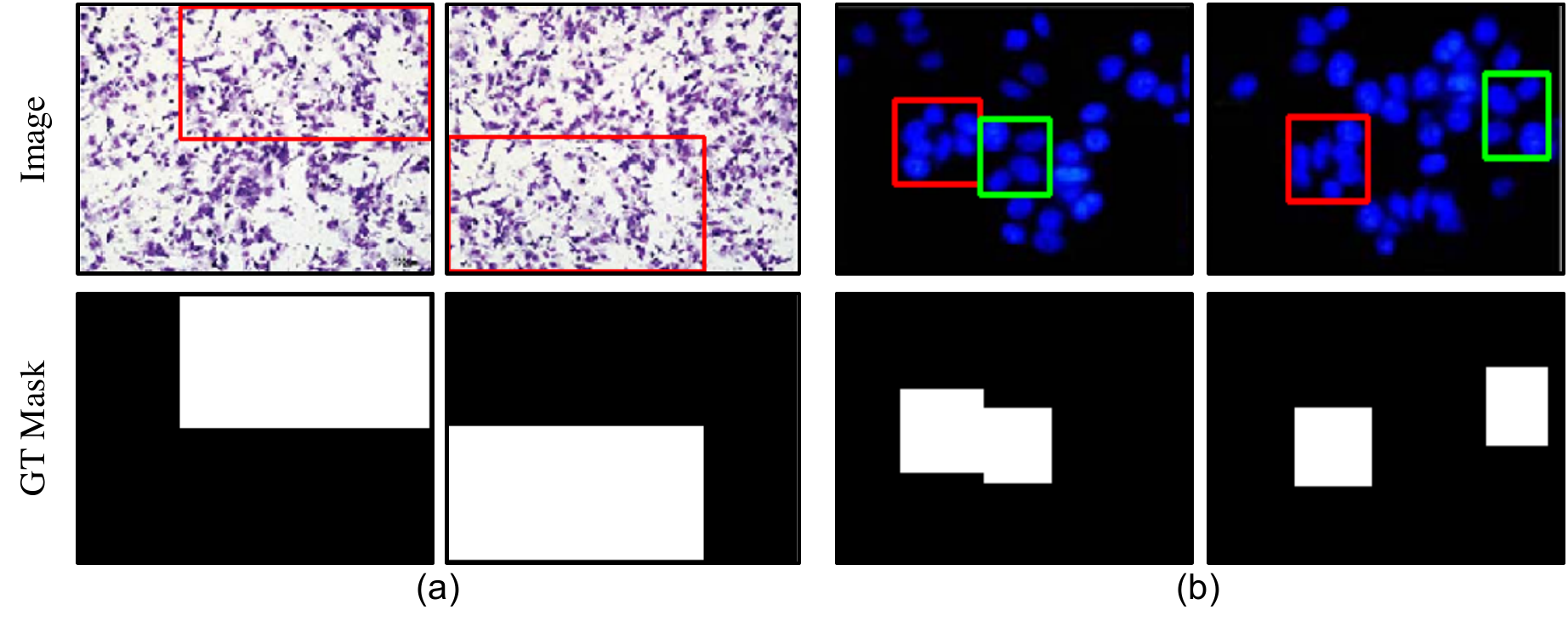}
    \caption{Two image pairs exhibiting duplication manipulation in EDD task. Duplicated regions are color coded to show correspondence. Bottom row shows ground truth masks for evaluation.}
    \label{duplication_annotation}
\end{figure}

\paragraph{External Duplication Detection (EDD):} This task involves detection of near identical regions between images. The duplicated region may span all or part of an image. Figure \ref{duplication_annotation} shows two examples of external duplication. Duplicated regions may appear due to two reasons --- (1) cropping two images with an overlap from a larger original source image and (2) by splicing i.e. copy-pasting a region from one image into another as shown in Figure \ref{duplication_annotation}a and b respectively. Irrespective of the origin of manipulation, the task requires detection of recurring regions between a pair of images. Further, another dimension of complexity for EDD stems from the orientation difference between duplicated regions. Duplicated regions in the second example of Figure \ref{duplication_annotation} have been rotated by $180^{\circ}$. We also found orientation difference of $0^{\circ}$, $90^{\circ}$, horizontal and vertical flip. From an evaluation perspective, an image pair is considered one sample for EDD task and ground truth masks also occur in pairs. The same image may have unique masks for different pairs corresponding to duplicated regions. Since, it is computationally expensive to consider all image pairs in a document, we drastically reduce the number of pairs to be computed by considering pairs of the same image class. This is a reasonable heuristic, since (1) we do not find duplications between images of different class and (2) automated image classification has reliable accuracy as shown in Table \ref{image_classification}. For statistics on orientation difference and more duplication examples please refer to the supplementary material.

\begin{figure}
    \centering
    \includegraphics[width=\linewidth]{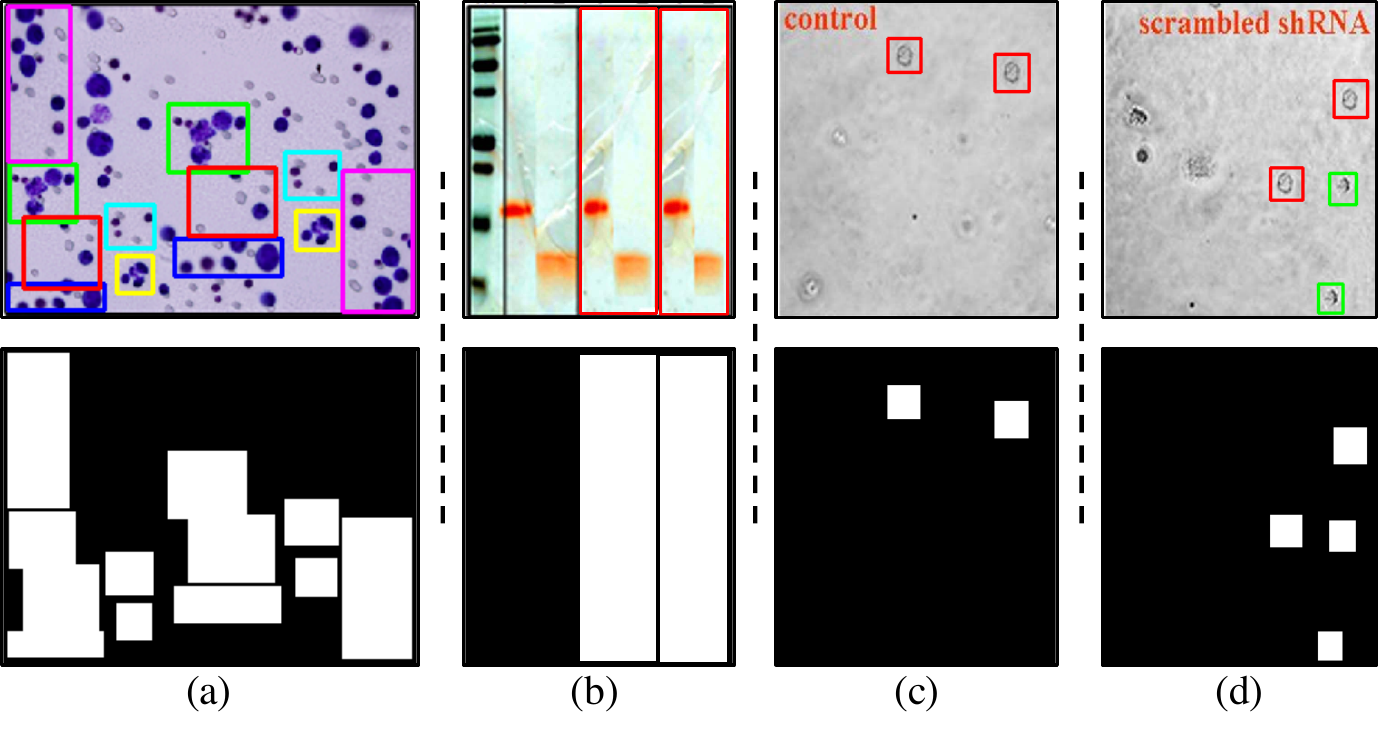}
    \caption{Manipulated samples in IDD task. Top row shows images and bottom row has corresponding masks. Repeated regions within the same image are color coded.}
    \label{copymove_annotation}
\end{figure}

\paragraph{Internal Duplication Detection (IDD):} IDD is our proposed image forensics task that involves detection of internally repeated image regions \cite{Wu_2018_ECCV,islam2020doa}. Unlike a standard copy-move forgery detection (CMFD) task where the source region is known and is also from the same image, in IDD the source region may or may not be from the same image. The repeated regions may have been procured by the manipulator from a different image or document. Figure \ref{copymove_annotation} shows examples of internal duplication. Notice that the regions highlighted in red in Figure \ref{copymove_annotation}c and d are the same and it is unclear which or if any of the patches is the source. Consequently from an evaluation perspective we treat all duplicated regions within an image as forged. Ground truth annotation includes one mask per image.

\begin{figure}
    \centering
    \includegraphics[width=\linewidth]{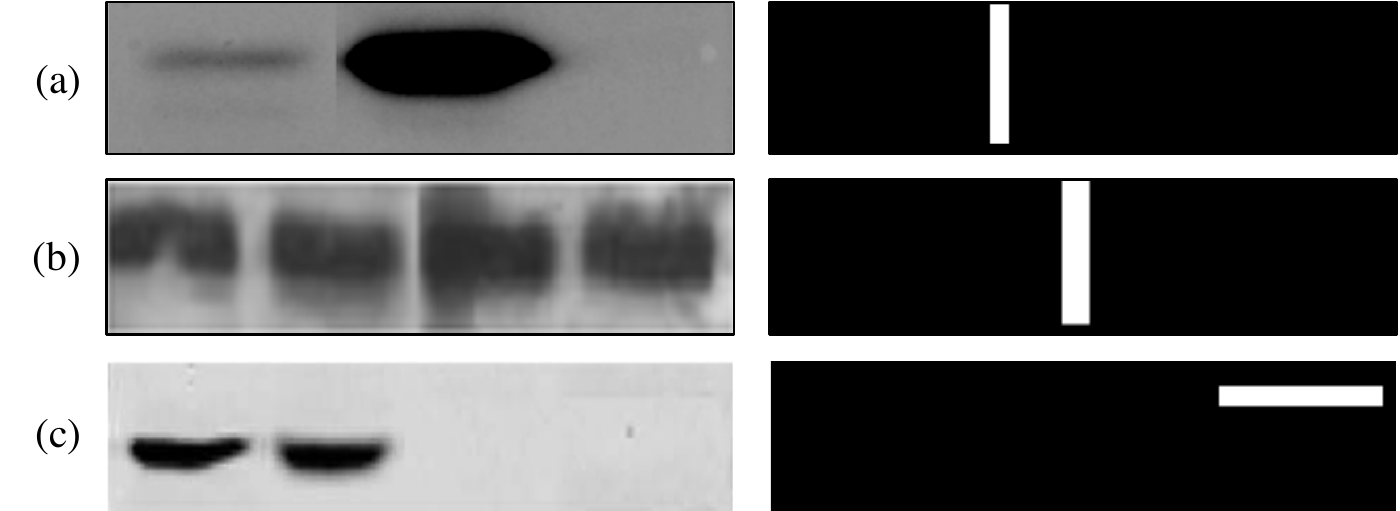}
    \caption{Examples of cuts/transitions. Noticeable sharp transition in (c) has been annotated, but the complete boundary is unclear.}
    \label{patchcut_annotation}
\end{figure}

\begin{figure}
    \centering
    \includegraphics[width=\linewidth]{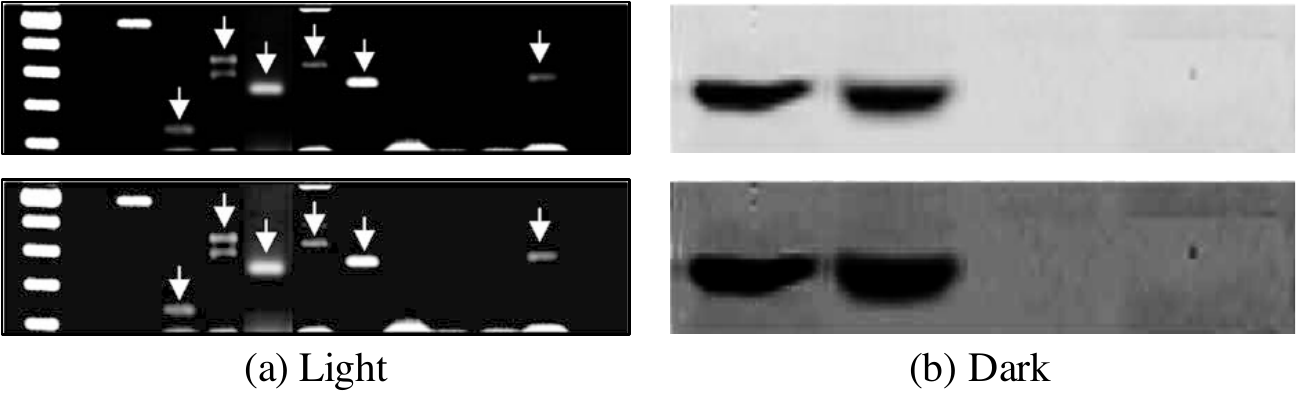}
    \caption{Left and right examples show light and dark gamma correction of images making it easier to spot potential manipulations. The third arrow band in (a) appears to be spliced.}
    \label{patchcut_exposure}
\end{figure}

\paragraph{Cut/Sharp-Transition Detection (CSTD):} A cut or a sharp transition can occur at the boundary of spliced or tampered regions. Unlike spliced images on social media, blot/gel images do not show a clear distinction between the authentic background and spliced foreground, making it difficult to identify the foreign patch. As an example, in Figure \ref{patchcut_annotation}a and b it is not possible to identify if the left or right section of the western blot is spliced. Sharp transitions in texture can also occur from blurring of pixels or other manipulations of unknown provenance. In both cases, a discontinuity in image-texture in the form of a cut or sharp transition is the sole clue to detect manipulations. Accordingly we annotate anomalous boundaries as forged. From an annotation perspective, cuts or sharp transitions can be difficult to see, therefore we used gamma correction to make the images light or dark and highlight manipulated regions. Figure \ref{patchcut_exposure} shows examples of gamma correction. Ground truth is a binary mask for each image.

\section{Why is Biomedical Forensics Hard?}\label{analysis}

Based on our insights from the data curation process and analysis of experimental results in Sec.~\ref{evaluation}, we explain potential challenges for natural-image forensic methods when applied to biomedical domain.

\noindent \textbf{Artifacts in Biomedical Images:}
\begin{figure}
    \centering
    \includegraphics[width=\linewidth]{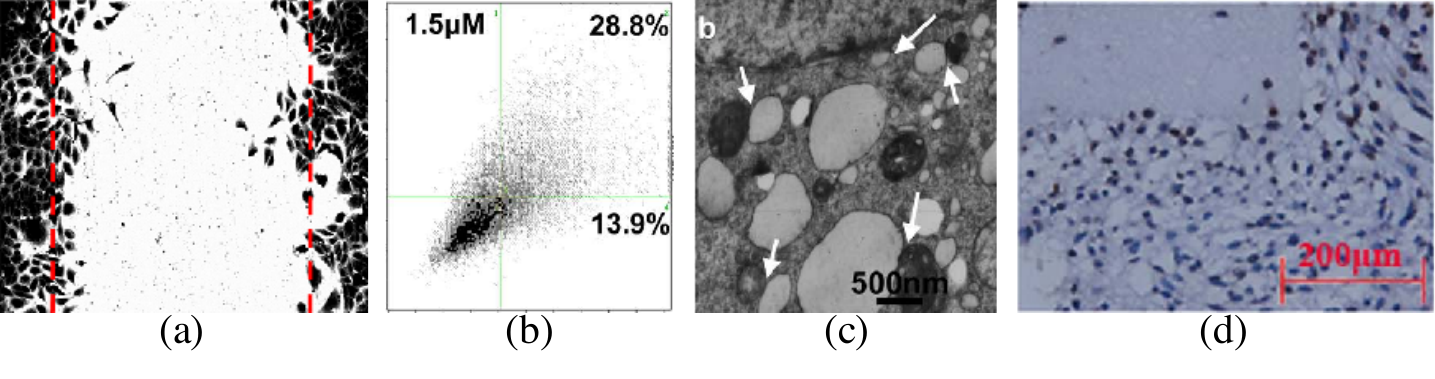}
    \caption{Examples of annotation artifacts in biomedical images: (a) dotted lines (b) alphanumeric text (c) arrows (d) scale.}
    \label{artifacts}
\end{figure}
Unlike natural image datasets, biomedical images are scientific images presented in research documents. Accordingly, there are artifacts in the form of annotations and legends that are added to an image. Figure \ref{artifacts} shows some common artifacts that we found, including text and symbols such as arrows, scale and lines. The presence of these artifacts can create false positive matches for EDD and IDD tasks.

\noindent \textbf{Figure Semantics:}
\begin{figure}
    \centering
    \includegraphics[width=\linewidth]{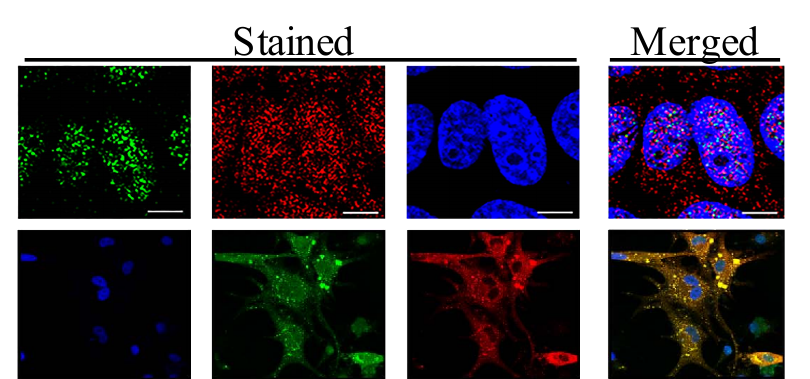}
    \caption{Left three columns show staining of microscopy images. Right column is an overlay of all stained images. Two or more images can be found tiled in this fashion.}
    \label{stains}
\end{figure}
\begin{figure}
    \centering
    \includegraphics[width=\linewidth]{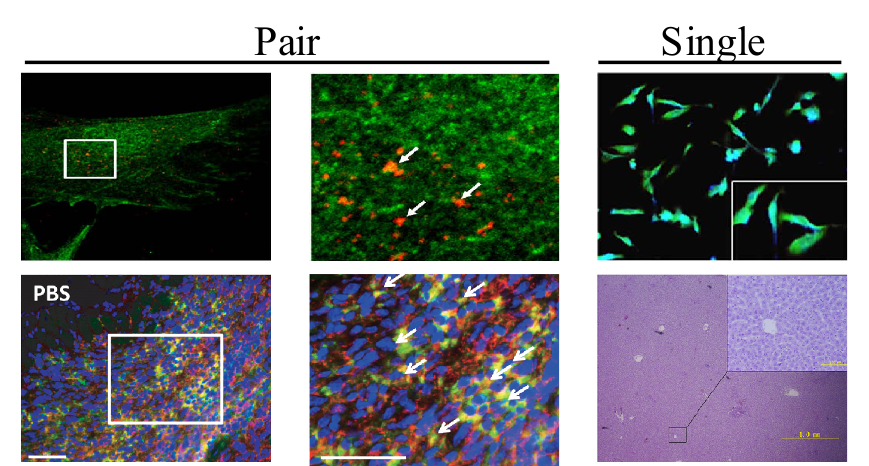}
    \caption{Images on the left show pairs of zoomed images. Right column has zoomed regions within the image. Rectangular bounding boxes are part of the original image.}
    \label{zoom}
\end{figure}
Biomedical research documents contain images that are visually similar, but the figure semantics indicates that they are not manipulated. Two such statistically significant semantics are staining-merging and zoom. Forgery detection algorithms may generate false positive matches for images belonging to these categories. Stained images originate from microscopy experiments that involve colorization of the same cell/tissue sample with different fluorescent chemicals. This is usually followed by a merged/overlaid image which combines the stained images. The resulting images are tiled together in the same figure. Since the underlying cell/tissue sample is unchanged, the image structure is retained across images but with color change.  Figure \ref{stains} shows some samples of staining and merging. The second semantics involves repeated portions of images that are magnified to highlight experimental results. Zoom semantics involves images that contain a zoomed portion of the image internally or themselves are a zoomed portion of another image. The zoomed area is indicated by a rectangular bounding box and images are adjacent. Figure \ref{zoom} shows paired and single images with zoom semantics. 

\begin{figure}[h]
    \centering
    \includegraphics[width=0.9\linewidth]{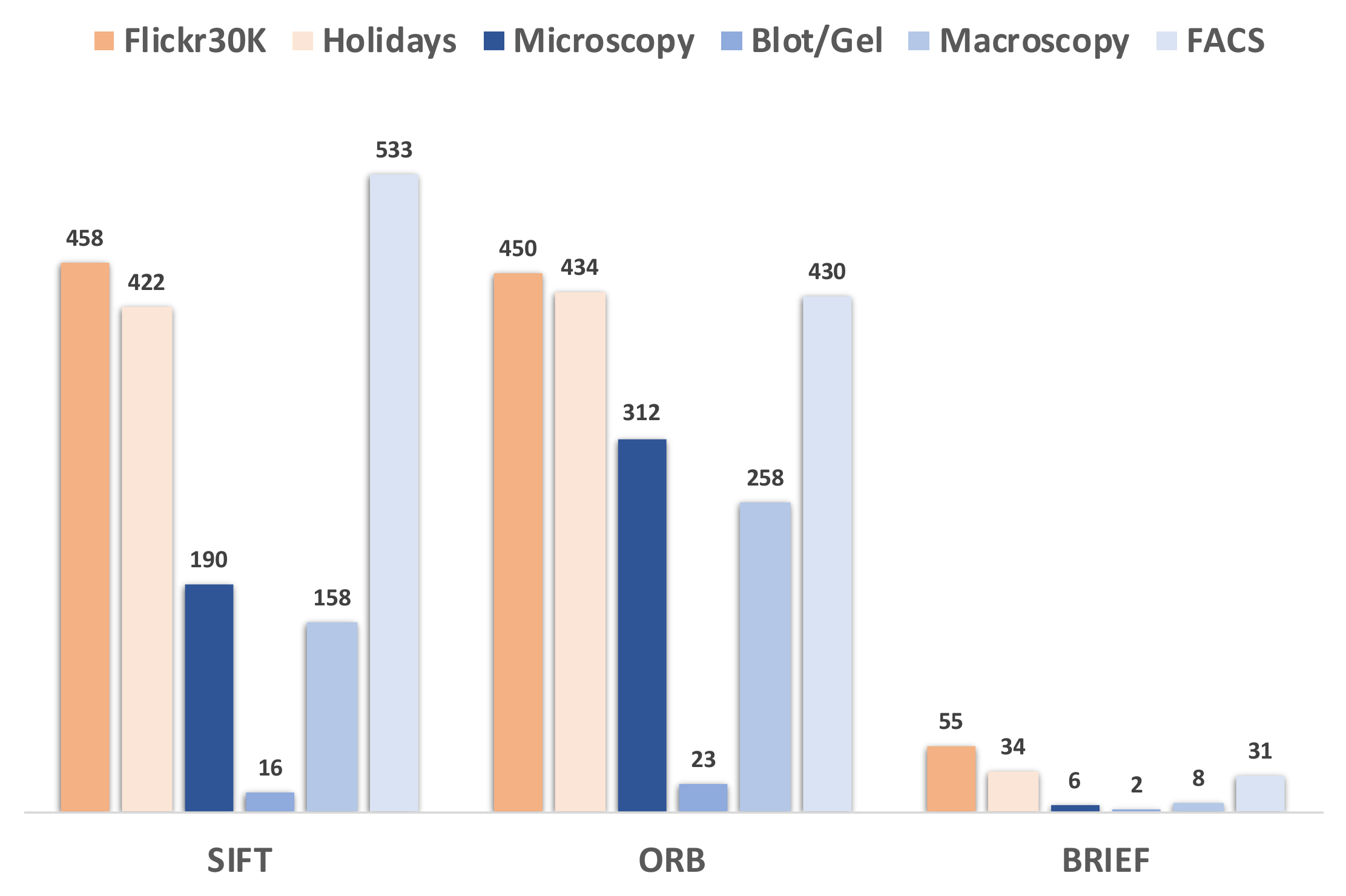}
    \caption{Median number of keypoints identified in images. Biomedical images have a relatively plain texture with the exception of FACS images, leading to fewer keypoints.}
    \label{texture}
\end{figure}

\noindent \textbf{Image Texture:}
As illustrated in Figure \ref{image_samples}, biomedical images tend to have a plain or pattern like texture with the exception of macroscopy images. This phenomena is particularly accentuated in blot/gel and microscopy images which are the largest two image classes and also contain the most manipulations. The plain texture of images makes it difficult to identify keypoints and extract descriptors for image matching, making descriptor based duplication detection difficult. We contrast this with the ease of identifying keypoints from two common computer vision datasets -- Flickr30k \cite{plummer2015flickr30k} and Holidays \cite{holidays}. Figure \ref{texture} shows the median number of keypoints identified in each image class using three off-the-shelf descriptor extractors: SIFT \cite{lowe2004distinctive}, ORB \cite{rublee2011orb}, BRIEF \cite{calonder2010brief}. We resized all images to 256x256 pixels to account for differing images sizes. With the exception of FACS, other three image classes show a sharp decline in the number of extracted keypoints. We consider FACS to be an exception due to the large number of dots, where each dot is capable of producing a keypoint. However these keypoints may be redundant and not necessarily useful for biomedical image forensics.

\noindent \textbf{Hard Negatives:}
\begin{figure}
    \centering
    \includegraphics[width=\linewidth]{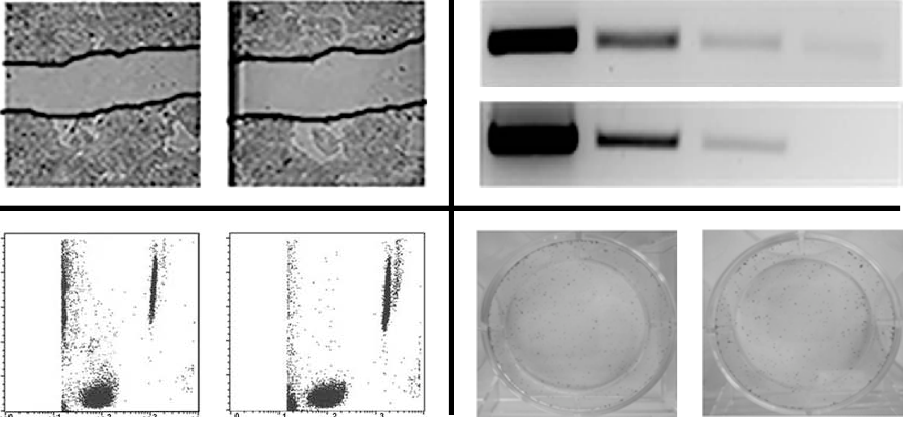}
    \caption{Hard negative samples from Blot/Gel, Macroscopy, FACS and Microscopy classes in clockwise order.}
    \label{hard_negative}
\end{figure}
Scientific experiments often involve tuning of multiple parameters in a common experimental paradigm to produce comparative results. For biomedical experiments, this can produce very similar-looking images, which can act like hard negatives when looking for duplicated regions. For blot and gel images this can be true irrespective of a common experimental framework due to patterns of blobs on a monotonous background. Figure \ref{hard_negative} shows some hard negative samples for each image class.

\section{Evaluation and Benchmarking}\label{evaluation}

\subsection{Metrics}
For all the manipulation tasks discussed in Section \ref{manipulation_tasks}, detection algorithms are expected to produce a binary prediction mask of the same dimension as the input image. The predicted masks are compared against ground truth annotation masks included in the dataset. Manipulated pixels in images denote the positive class. Following previous work in forgery detection \cite{Wu_2018_ECCV,Wu_2019_CVPR,wu2017deep} we compute $F_{1}$ scores between the predicted and ground truth mask for all tasks. We also compute Matthews correlation coefficient (MCC) \cite{matthews1975comparison} between the masks since it has been shown  to present a balanced score when dealing with imbalanced data \cite{chicco2020advantages,boughorbel2017optimal} as is our case with fewer manipulated images. MCC ranges from -1 to +1 and represents the correlation between prediction and ground truth. Due to space constraints, $F_{1}$ score tabulation is done in supplementary material. Evaluation is done both at the image and pixel-level i.e. true/false positives and true/false negatives are determined for each image and pixel. For image evaluation, following the protocol in \cite{Wu_2018_ECCV}, we consider an image to be manipulated if any one pixel has positive prediction. Pixel level evaluation across multiple images is similar to protocol A in \cite{Wu_2018_ECCV} i.e. all pixels from the dataset are gathered for one final computation.

\subsection{Baseline Models}

We evaluate several deep learning and non deep learning models for our three tasks introduced in Section \ref{manipulation_tasks}. Our baselines are selected from forensics literature based on model/code availability and task suitability. Deep-learning baselines require finetuning for weight adaptation. However, due to small number of manipulated samples, BioFors training set comprises pristine images only. Inspired by previous forgery detection methods \cite{Wu_2018_ECCV,wu2017deep}, we create synthetic manipulations on pristine training data to finetune models. Details of synthetic data and baseline experiments are provided in the supplementary material. To promote reproducibility, our synthetic data generators and evaluation scripts will be released with the dataset. 

\begin{table*}[t]
\small
    \centering
    \begin{tabular}{lcccccccccc}
    \toprule
         \multirow{3}{*}{\textbf{Method}} &  \multicolumn{2}{c}{\textbf{Microscopy}} & \multicolumn{2}{c}{\textbf{Blot/Gel}} & \multicolumn{2}{c}{\textbf{Macroscopy}} & \multicolumn{2}{c}{\textbf{FACS}} & \multicolumn{2}{c}{\textbf{Combined}} \\
         \cmidrule(lr){2-3}\cmidrule(lr){4-5}\cmidrule(lr){6-7}\cmidrule(lr){8-9}\cmidrule(lr){10-11}
         & Image & Pixel & Image & Pixel & Image & Pixel & Image & Pixel & Image & Pixel \\
         \cmidrule(r){1-1}\cmidrule(l){2-11}
         SIFT \cite{lowe2004distinctive} & 0.180 & 0.146 & 0.113 & 0.148 & 0.130 & 0.194 & 0.11 & 0.073 & 0.142 & 0.132 \\
         ORB \cite{rublee2011orb} & 0.319 & 0.342 & 0.087 & 0.127 & 0.126 & 0.226 & 0.269 & 0.187 & 0.207 & 0.252 \\
         BRIEF \cite{calonder2010brief} & 0.275 & 0.277 & 0.058 & 0.102 & 0.135 & 0.169 & 0.244 & 0.188 & 0.180 & 0.202 \\
         DF - ZM \cite{cozzolino2015efficient} & \textbf{0.422} & \textbf{0.425} & 0.161 & 0.192 & \textbf{0.285} & \textbf{0.256} & \textbf{0.540} & \textbf{0.504} & \textbf{0.278} & \textbf{0.324} \\
         DMVN \cite{wu2017deep} & 0.242 & 0.342 & \textbf{0.261} & \textbf{0.430} & 0.185 & 0.238 & 0.164 & 0.282 & 0.244 & 0.310 \\
         \bottomrule
    \end{tabular}
    \caption{Results for external duplication detection (EDD) task by image class. Image and Pixel columns denote image and pixel level evaluation respectively. All numbers are MCC scores. For corresponding $F_{1}$ scores, please refer to supplementary material.}
    \label{duplication_detection}
\end{table*}

\begin{table*}[!htbp]
\small
    \centering
    \begin{tabular}{lcccccccc}
    \toprule
         \multirow{3}{*}{\textbf{Method}} &  \multicolumn{2}{c}{\textbf{Microscopy}} & \multicolumn{2}{c}{\textbf{Blot/Gel}} & \multicolumn{2}{c}{\textbf{Macroscopy}} & \multicolumn{2}{c}{\textbf{Combined}} \\
         \cmidrule(lr){2-3}\cmidrule(lr){4-5}\cmidrule(lr){6-7}\cmidrule(lr){8-9}
         & Image & Pixel & Image & Pixel & Image & Pixel & Image & Pixel \\
         \cmidrule(r){1-1}\cmidrule(l){2-9}
         DF - ZM \cite{cozzolino2015efficient} & \textbf{0.764} & 0.197 & \textbf{0.515} & 0.449 & 0.573 & 0.478 & 0.564 & 0.353 \\
         DF - PCT \cite{cozzolino2015efficient} & \textbf{0.764} & \textbf{0.202} & 0.503 & \textbf{0.466} & \textbf{0.712} & \textbf{0.487} & \textbf{0.569} & \textbf{0.364} \\
         DF - FMT \cite{cozzolino2015efficient} & 0.638 & 0.167 & 0.480 & 0.400 & 0.495 & 0.458 & 0.509 & 0.316 \\
         DCT \cite{fridrich2003detection} & 0.187 & 0.022 & 0.250 & 0.168 & 0.158 & 0.143 & 0.196 & 0.095 \\
         DWT \cite{dwt} & 0.299 & 0.067 & 0.384 & 0.295 & 0.591 & 0.268 & 0.341 & 0.171 \\
         Zernike \cite{ryu2010detection} & 0.192 & 0.032 & 0.336 & 0.187 & 0.493 & 0.262 & 0.257 & 0.114 \\
         BusterNet \cite{Wu_2018_ECCV} & 0.183 & 0.178 & 0.226 & 0.076 & 0.021 & 0.106 & 0.269 & 0.107 \\
         \bottomrule
    \end{tabular}
    \caption{Results for internal duplication detection (IDD) task by image class and a combined result. There are no IDD instances in FACS images. Image and Pixel columns denote image and pixel level evaluation respectively. All numbers are MCC scores.}
    \label{copymove_detection}
\end{table*}

\paragraph{External Duplication Detection (EDD):} Baselines for EDD should identify repeated regions between images. We evaluate classic keypoint-descriptor based image-matching algorithms such as SIFT \cite{lowe2004distinctive}, ORB \cite{rublee2011orb} and BRIEF \cite{calonder2010brief}. We follow a classic object matching approach, using RANSAC \cite{fischler1981random} to remove stray matches. CMFD algorithms can be used by concatenating two images to create a single input. We evaluated DenseField (DF) \cite{cozzolino2015efficient} with best reported transform -- zernike moment (ZM) on concatenated images. Additionally, we evaluate a splicing detection algorithm, DMVN \cite{wu2017deep} to find repeated regions. DMVN implements a deep feature correlation layer which matches coarse image features at 16x16 resolution to find visually similar regions.

\paragraph{Internal Duplication Detection (IDD):} Appropriate baselines for IDD should be suitable for identifying repeated regions within images. DenseField (DF) \cite{cozzolino2015efficient} proposes an efficient dense feature matching algorithm for CMFD. We evaluate it using the three circular harmonic transforms used in the paper: zernike moments (ZM), polar cosine transform (PCT) and fourier-mellin transform (FMT). We also evaluated the CMFD algorithm reported in \cite{christlein2012evaluation}, using three block based features -- discrete cosine transform (DCT) \cite{fridrich2003detection}, zernike moments (ZM) \cite{ryu2010detection}  and discrete wavelet transform (DWT) \cite{dwt}. BusterNet \cite{Wu_2018_ECCV} is a two-stream deep-learning based CMFD model that leverages visual similarity and manipulation artifacts. Visual similarity in BusterNet is identified using a self-correlation layer on coarse image features followed by percentile pooling.

\paragraph{Cut/Sharp-Transition Detection (CSTD):} Unlike the previous two tasks, it is challenging to find forensics algorithms designed for detecting  cuts or transitions. We evaluate ManTraNet \cite{Wu_2019_CVPR}, a state-of-the-art manipulation detection algorithm which identifies anomalous pixels and image regions. We also evaluated a baseline convolutional neural network (CNN)  model for detecting cuts and transitions. The CNN was trained on synthetic manipulations in blot/gel images from the training set. For more details on the baseline please refer to supplementary material.

\subsection{Results}

\begin{table}[]
\small
    \centering
    \begin{tabular}{lrrrr}
    \toprule
         \multirow{2}{*}{\textbf{Method}} & \multicolumn{2}{c}{$F_{1}$} & \multicolumn{2}{c}{MCC} \\
         \cmidrule(lr){2-3}\cmidrule(lr){4-5}
         & Image & Pixel & Image & Pixel \\
         \cmidrule(r){1-1}\cmidrule(l){2-5}
         MantraNet \cite{Wu_2019_CVPR} & \textbf{0.253} & \textbf{0.09} & \textbf{0.170} & \textbf{0.080} \\
         CNN Baseline & 0.212 & 0.08 & 0.098 & 0.070 \\
         \bottomrule
    \end{tabular}
    \caption{Results on the cut/sharp-transition detection (CSTD) task.}
    \label{patch_cut_detection}
\end{table}

Tables \ref{duplication_detection},~\ref{copymove_detection} and \ref{patch_cut_detection} present baseline results for EDD, IDD and CSTD tasks respectively. We find that dense feature matching approaches (DF-ZM,PCT,FMT) are better than sparse (SIFT, SURF, ORB), block-based (DCT, DWT, Zernike) or coarse feature matching methods (DMVN and BusterNet) for identifying repeated regions in both EDD and IDD tasks. Dense feature matching is computationally expensive, and most image forensics algorithms obtain a viable quality-computation trade-off on natural images. However, biomedical images have relatively plain texture and similar patterns, which may lead to indistinguishable features for coarse or sparse extraction. For the set of baselines evaluated, exchanging feature matching quality for computation is not successful on biomedical images. Furthermore, performance varies drastically across image classes for all methods, with models peaking across different image classes. The variation is expected since the semantic and visual characteristics vary by image category. However, as a direct consequence of this variance, image category specific models may need to be developed in future research. On CSTD, our simple baseline trained to detect sharp transitions produces false alarms on image borders or edges of blots. Both MantraNet and our baseline have similar performance, indicating that a specialized model design might be required to detect cuts and anomalous transitions. Finally, performance is low across all tasks which can be attributed to some of the challenges discussed in Section \ref{analysis}. In summary, it is safe to conclude that existing natural-image forensic methods are not robust when applied to biomedical images and also show high variation in performance across image classes. The results emphasize the need for robust forgery detection algorithms that are applicable to the biomedical domain. For sample predictions from reported baselines please refer to the supplementary material.
\section{Conclusion and Future Work}

Manipulation of scientific images is an issue of serious concern for the biomedical community. While reviewers can attempt to screen for scientific misconduct, the complexity and volume of the task places an undue burden on them. Automated and scalable biomedical forensic methods are necessary to assist reviewers. We presented BioFors, a large biomedical image forensics dataset. BioFors comprises a comprehensive range of images found in biomedical documents. We also framed three manipulation detection tasks based on common manipulations found in literature. Our evaluations show that common computer vision algorithms are not robust when extended to the biomedical domain. Our analysis shows that attaining respectable performance will require well designed models, as there are multiple challenges to the problem. We expect that BioFors will advance biomedical image forensic research.

\section{Acknowledgement}

We are deeply grateful to Dr. Elisabeth M. Bik, Dr. Arturo Casadevall and Dr. Ferric C. Fang for sharing with us the raw annotation of manipulations. Their contributions accelerated the creation of the dataset we have released. We also extend our special thanks to Dr. Bik for answering many of our questions and improving our understanding of the domain of biomedical images.

\clearpage


{\small
\bibliographystyle{ieee_fullname}
\bibliography{egbib}
}

\clearpage


\begin{center}
    \centering
    \LARGE
    \title{\textbf{Supplementary Material}}\\
\end{center}

\maketitle
\appendix
\section{Image Collection and Statistics}

\begin{figure}[!h]
    \centering
    \includegraphics[width=\linewidth]{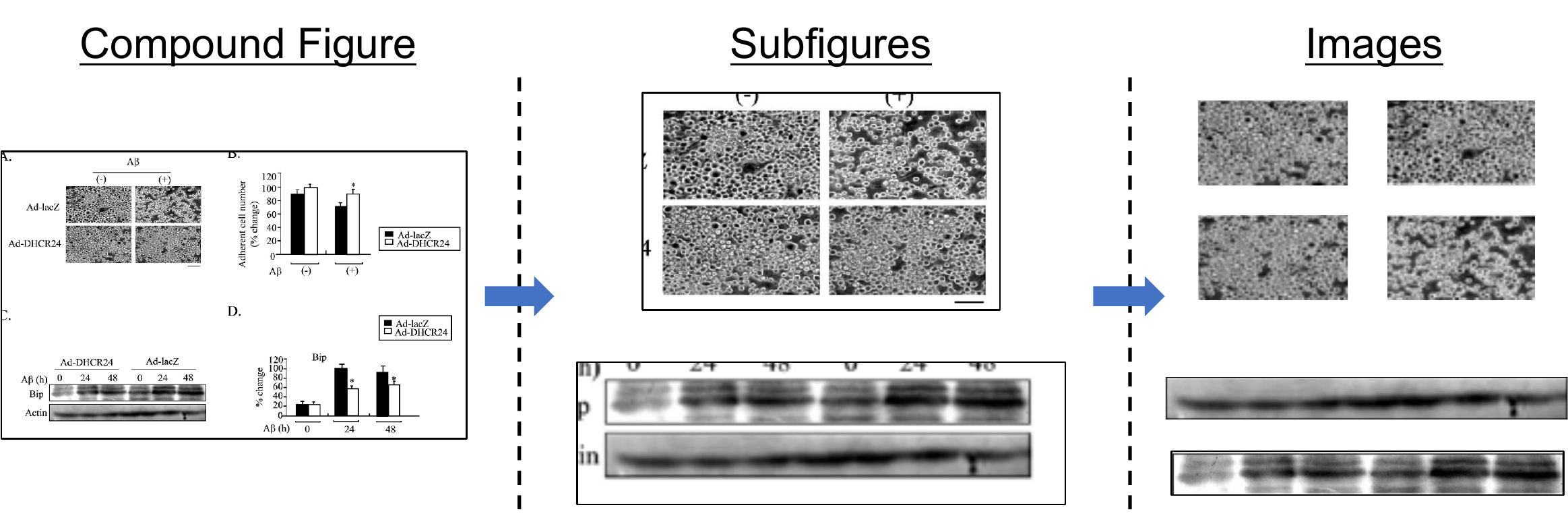}
    \caption{We crop compound biomedical figures in two stages: 1) crop sub-figures and 2) crop images from sub-figures. Synthetic images such as charts and plots are filtered.}
    \label{image_collection}
\end{figure}

\begin{figure}[!h]
    \centering
    \includegraphics[width=\linewidth]{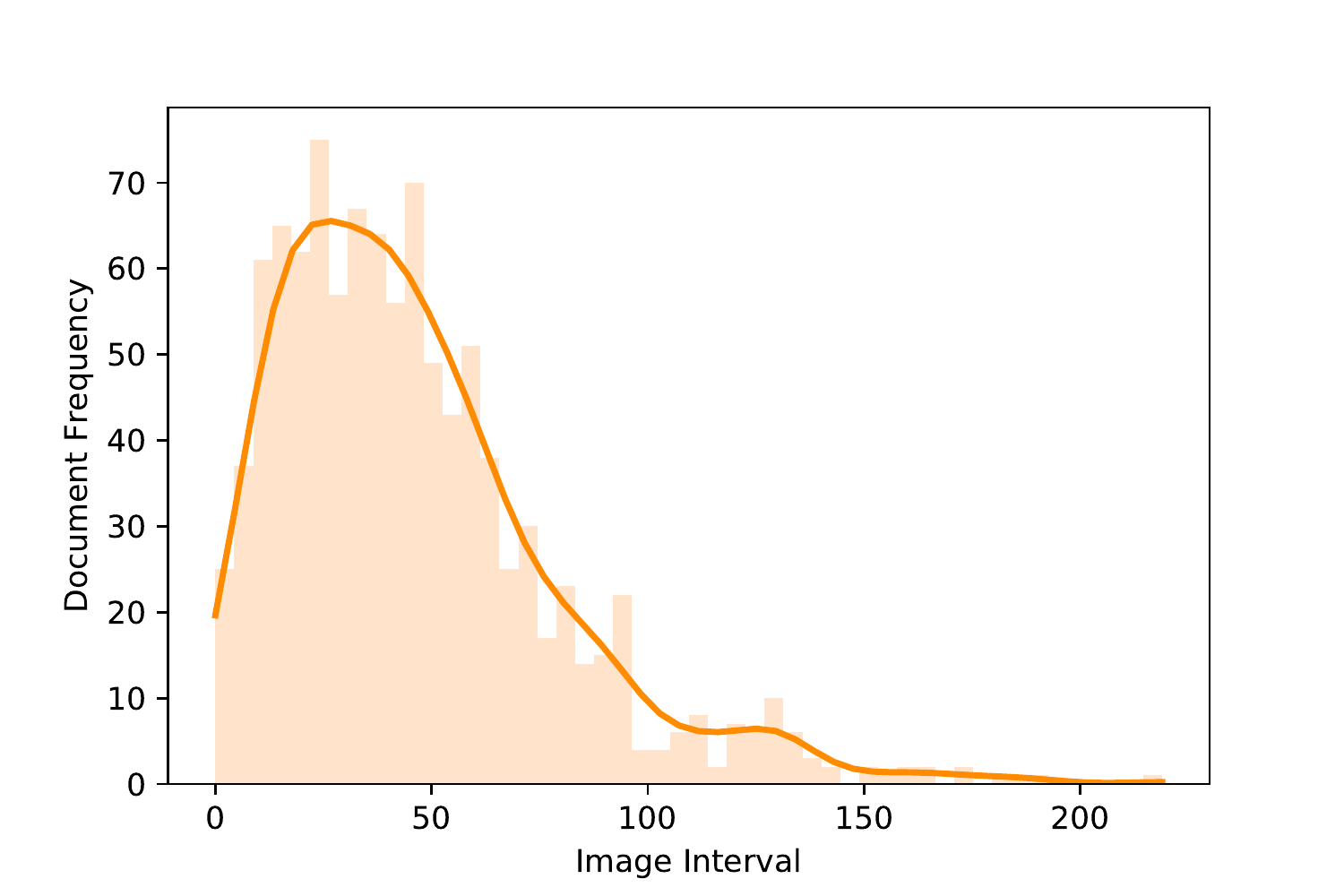}
    \caption{Distribution of images extracted from documents. The distribution peaks at 25 images from most documents. The rightmost entry has 219 images from one document.}
    \label{image_dist}
\end{figure}

\begin{figure}[!h]
    \centering
    \includegraphics[width=\linewidth]{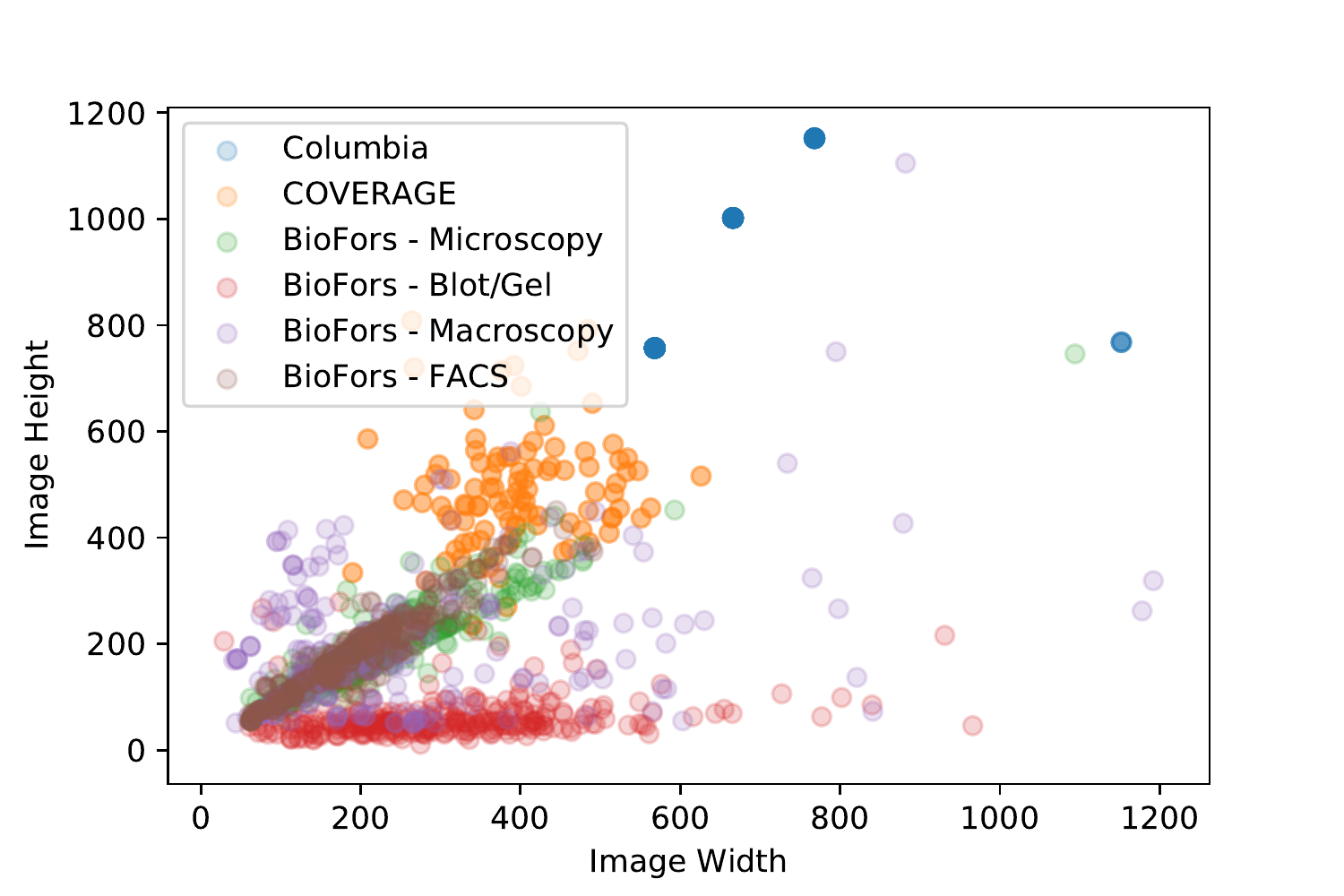}
    \caption{BioFors images have much higher variation in dimension compared to two popular image forensic datasets.}
    \label{image_areas}
\end{figure}

We described the image collection procedure from compound biomedical figures in the main paper. The overall procedure involved automatic extraction of compound figures from documents followed by  manual cropping of images. Figure \ref{image_collection} shows a sample compound figure and its decomposition. Furthermore, there is a significant variation in the number of images extracted from each document. Figure \ref{image_dist} shows the frequency of images extracted per document. Finally, images in BioFors have a wide range of dimensions. Figure \ref{image_areas} shows a scatter-plot of BioFors image dimensions as compared to two other natural-image forensic datasets, Columbia \cite{ng2009columbia} and COVERAGE \cite{wen2016coverage}.

\section{Orientation}

\begin{figure}
    \centering
    \includegraphics[width=0.93\linewidth]{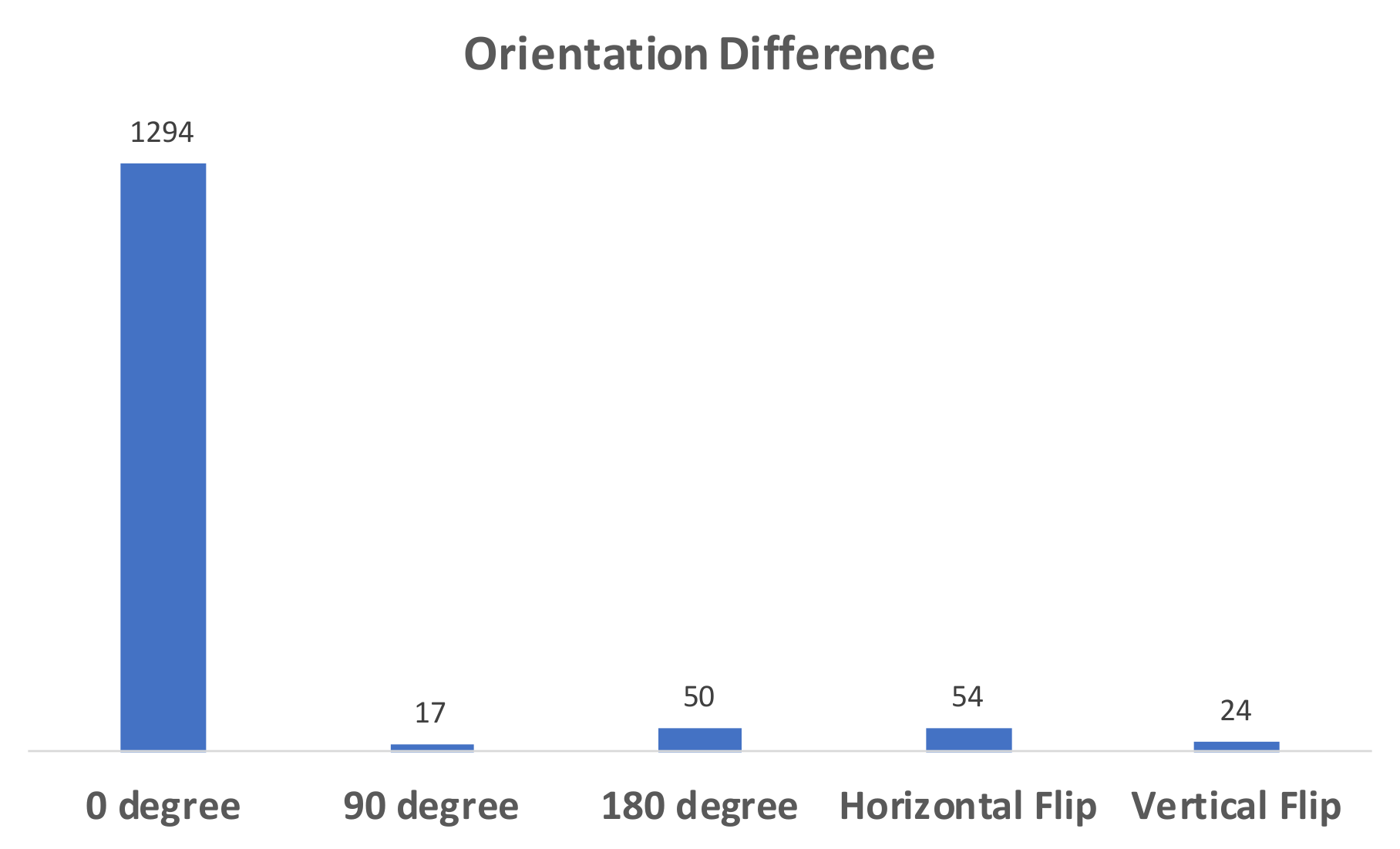}
    \caption{Frequency of differing orientations between duplicated regions in EDD and IDD tasks.}
    \label{orientation_diff}
\end{figure}

Duplicated regions in BioFors may have an orientation difference. These differences may occur between duplicated regions across two images (external duplication detection (EDD) task) or within an image (internal duplication detection (IDD) task). We found five major categories of differing orientation: $0^{\circ}$, $90^{\circ}$, $180^{\circ}$, horizontal and vertical flip. Figure \ref{orientation_diff} shows the frequency of each orientation between duplicated regions.

\section{Baseline for CSTD}

\begin{figure}
    \centering
    \includegraphics[width=0.93\linewidth]{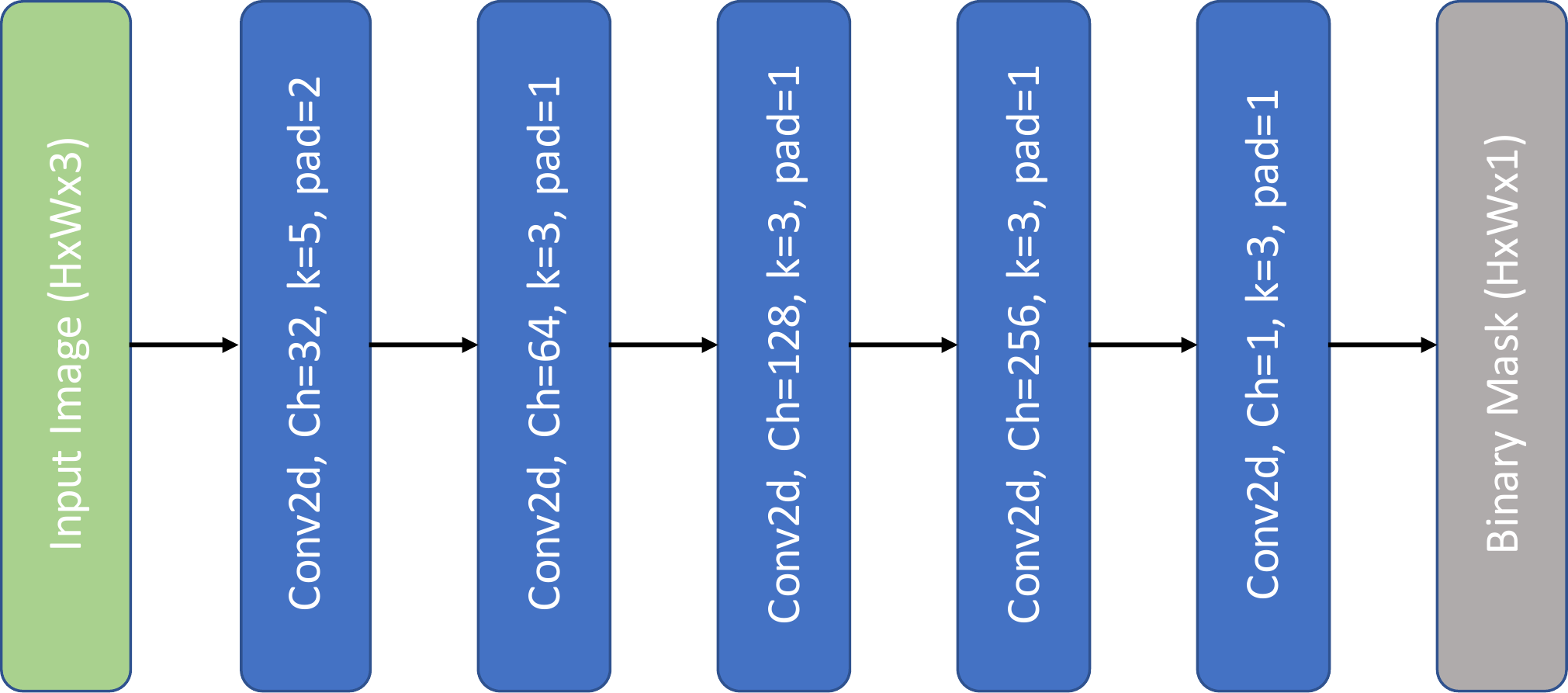}
    \caption{Our baseline CNN architecture. Padding ensures the same height and width for input image and output mask.}
    \label{cnn_baseline}
\end{figure}

Existing forgery detection methods are not specifically designed to detect cuts/sharp transitions in images. The absence of diverse detection methods prompted us to train a simple convolutional neural network (CNN) baseline trained on synthetic manipulations in pristine blot/gel images from BioFors. Figure \ref{cnn_baseline} shows the CNN architecture. 

\section{Synthetic Data Generation}

Image forensic datasets usually do not have sufficient samples to train deep learning models. Previous works \cite{wu2017deep,Wu_2018_ECCV} created suitable synthetic manipulations in natural images for model pre-training. The synthetic manipulations were created by extracting objects from images and pasting them in the target image with limited data augmentation such as rotation and scale change. Similar to previous works, we created suitable synthetic manipulations in biomedical images corresponding to each task for training and validation. However, biomedical images do not have well defined objects and boundaries and manipulated regions are created with rectangular patches. Manipulation process for each task is discussed ahead.

\begin{figure}
    \centering
    \includegraphics[width=0.97\linewidth]{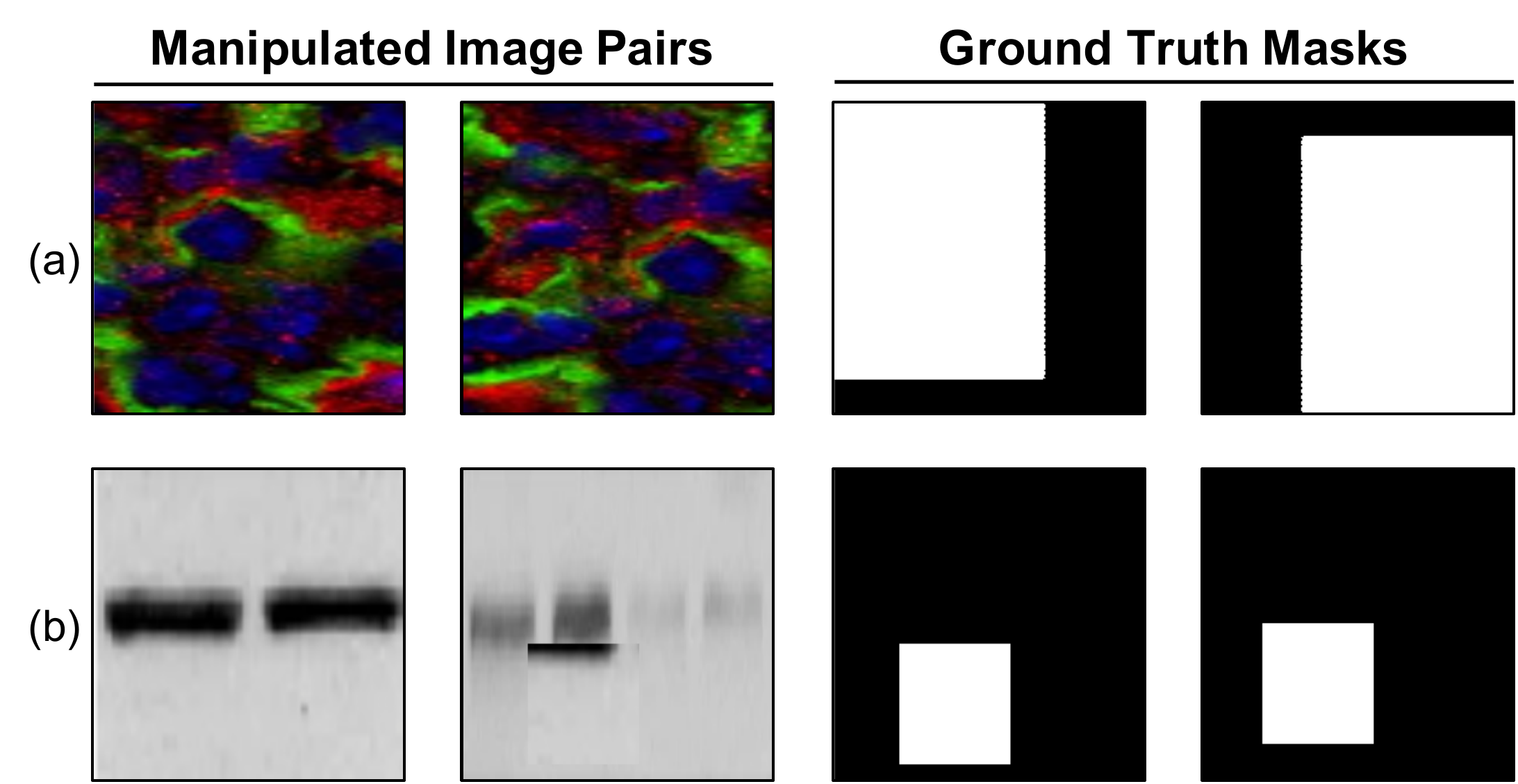}
    \caption{Synthetic manipulations in image pairs created using (a) overlapping image regions and (b) spliced image patches.}
    \label{edd_synth}
\end{figure}

\noindent \textbf{External Duplication Detection (EDD):}
Corresponding to the two possible sources of external duplication, we create manipulations by 1) Cropping two images with overlapping regions from a source image and 2) Splicing i.e. copy-pasting rectangular patches from source to target image. Manipulations of both types are shown in Figure \ref{edd_synth}. We generate pristine, spliced and overlapped samples in a 1:1:1 ratio. Images extracted with overlapping regions are resized to 256x256 image dimensions.

\begin{figure}
    \centering
    \includegraphics[width=0.94\linewidth]{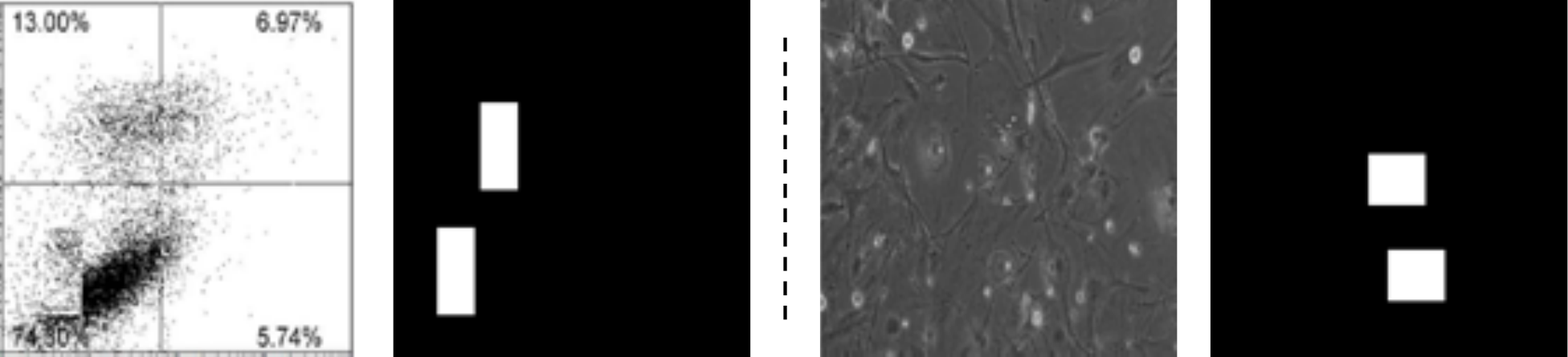}
    \caption{Internal duplications created with copy-move operations.}
    \label{idd_synth}
\end{figure}

\noindent \textbf{Internal Duplication Detection (IDD):} Internal duplications are created with a copy-move operation within an image. Rectangular patches of random dimensions are copy-pasted within the same image. Figure \ref{idd_synth} shows samples. 

\begin{figure}
    \centering
    \includegraphics[width=0.94\linewidth]{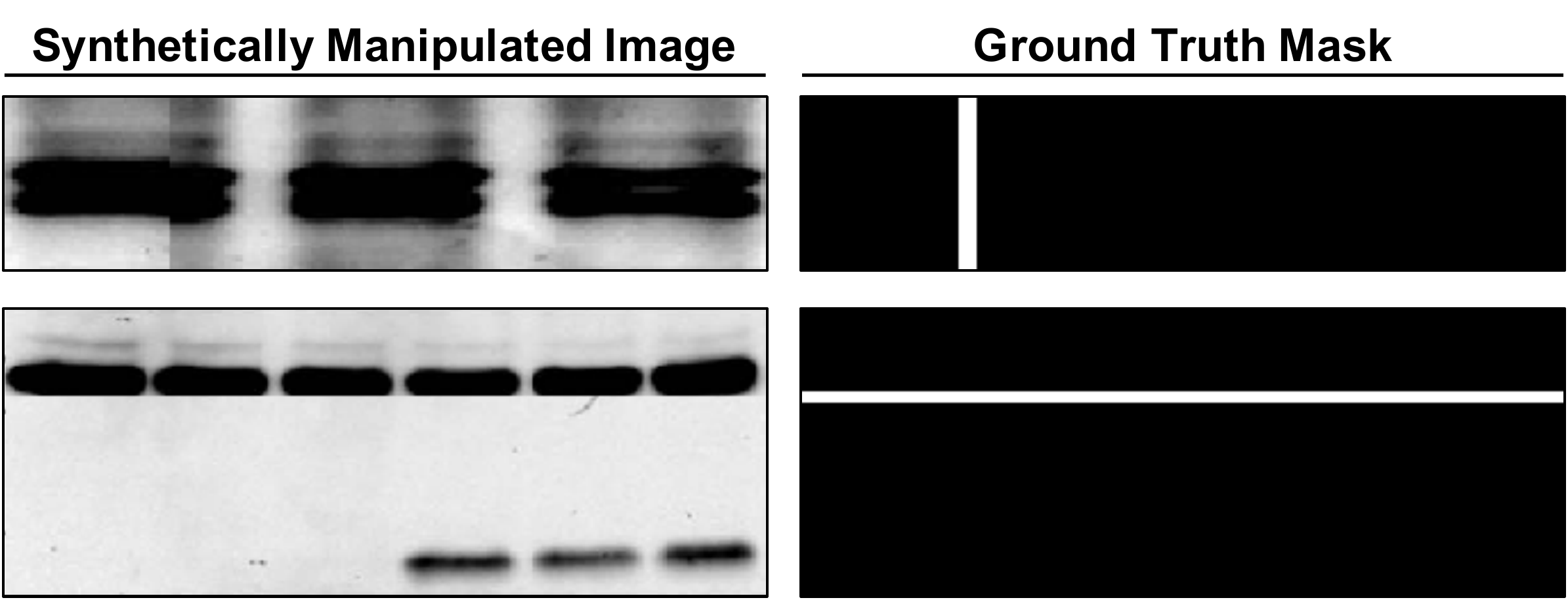}
    \caption{Synthetic cut/sharp transitions created in blot/gel images.}
    \label{cstd_synth}
\end{figure}

\noindent \textbf{Cut/Sharp-Transition Detection (CSTD):} We simulated synthetic manipulations by randomly splitting an image along two horizontal or vertical lines in the image and rejoining the image. The line of rejoining represents a synthetic cut or sharp transition and is used for training. Figure \ref{cstd_synth} shows synthetic CSTD manipulations.

\section{Experiment Details}

\begin{table*}[!t]
\small
    \centering
    \begin{tabular}{lcccccccccc}
    \toprule
         \multirow{3}{*}{\textbf{Method}} &  \multicolumn{2}{c}{\textbf{Microscopy}} & \multicolumn{2}{c}{\textbf{Blot/Gel}} & \multicolumn{2}{c}{\textbf{Macroscopy}} & \multicolumn{2}{c}{\textbf{FACS}} & \multicolumn{2}{c}{\textbf{Combined}} \\
         \cmidrule(lr){2-3}\cmidrule(lr){4-5}\cmidrule(lr){6-7}\cmidrule(lr){8-9}\cmidrule(lr){10-11}
         & Image & Pixel & Image & Pixel & Image & Pixel & Image & Pixel & Image & Pixel \\
         \cmidrule(r){1-1}\cmidrule(l){2-11}
         SIFT \cite{lowe2004distinctive} & 8.48\% & 5.82\% & 9.37\% & 11.74\% & 6.98\% & 10.52\% & 6.09\% & 2.32\% & 8.18\% & 5.83\% \\
         ORB \cite{rublee2011orb} & 30.48\% & 28.56\% & 5.97\% & 12.45\% & 9.87\% & 19.34\% & 22.53\% & 8.86\% & 20.66\% & 21.47\% \\
         BRIEF \cite{calonder2010brief} & 27.42\% & 25.59\% & 3.74\% & 9.74\% & 13.07\% & 16.22\% & 20.09\% & 9.33\% & 18.22\% & 18.58\% \\
         DF - ZM \cite{cozzolino2015efficient} & \textbf{42.00\%} & \textbf{42.08\%} & 15.42\% & 19.17\% & \textbf{27.48\%} & \textbf{25.82\%} & \textbf{54.17\%} & \textbf{50.24\%} & \textbf{27.06\%} & \textbf{32.46\%} \\
         DMVN \cite{wu2017deep} & 16.40\% & 31.54\% & \textbf{18.61\%} & \textbf{42.07\%} & 10.29\% & 21.78\% & 8.94\% & 18.85\% & 16.31\% & 27.55\% \\
         \bottomrule
    \end{tabular}
    \caption{$F_{1}$ scores for external duplication detection (EDD) task. The scores correspond to the experiments in main document.}
    \label{duplication_detection}
\end{table*}

\begin{table*}[!htbp]
\small
    \centering
    \begin{tabular}{lcccccccc}
    \toprule
         \multirow{3}{*}{\textbf{Method}} &  \multicolumn{2}{c}{\textbf{Microscopy}} & \multicolumn{2}{c}{\textbf{Blot/Gel}} & \multicolumn{2}{c}{\textbf{Macroscopy}} & \multicolumn{2}{c}{\textbf{Combined}} \\
         \cmidrule(lr){2-3}\cmidrule(lr){4-5}\cmidrule(lr){6-7}\cmidrule(lr){8-9}
         & Image & Pixel & Image & Pixel & Image & Pixel & Image & Pixel \\
         \cmidrule(r){1-1}\cmidrule(l){2-9}
         DF - ZM \cite{cozzolino2015efficient} & \textbf{74.07\%} & 12.04\% & \textbf{52.46\%} & 43.99\% & 53.33\% & 38.95\% & 56.10\% & 30.48\% \\
         DF - PCT \cite{cozzolino2015efficient} & \textbf{74.07\%} & \textbf{12.45\%} & 51.97\% & \textbf{46.02\%} & \textbf{70.59\%} & \textbf{40.03\%} & \textbf{57.31\%} & \textbf{31.90\%} \\
         DF - FMT \cite{cozzolino2015efficient} & 58.33\% & 9.94\% & 50.00\% & 38.53\% & 49.18\% & 39.84\% & 50.62\% & 27.74\% \\
         DCT \cite{fridrich2003detection} & 16.33\% & 3.50\% & 28.57\% & 17.06\% & 23.53\% & 15.53\% & 23.39\% & 10.35\% \\
         DWT \cite{dwt} & 25.97\% & 7.53\% & 40.00\% & 25.73\% & 63.16\% & 22.16\% & 37.04\% & 16.23\% \\
         Zernike \cite{ryu2010detection} & 17.95\% & 3.94\% & 34.78\% & 18.45\% & 42.86\% & 13.49\% & 28.99\% & 11.13\% \\
         BusterNet \cite{Wu_2018_ECCV} & 13.99\% & 13.38\% & 27.33\% & 4.57\% & 24.14\% & 14.94\% & 30.25\% & 6.04\% \\
         \bottomrule
    \end{tabular}
    \caption{$F_{1}$ scores for external duplication detection (EDD) task. The scores correspond to the experiments in main document.}
    \label{copymove_detection}
\end{table*}

We list the hyper-parameter and finetuning details of baselines corresponding to each task.

\noindent \textbf{Keypoint-Descriptor:} We implemented classic image matching algorithm using keypoint-descriptor based methods such as SIFT, ORB and BRIEF. Keypoints are matched using kd-tree and a consistent homography is found using RANSAC to remove outlier matches. A rectangular bounding box is created around the furthest matched keypoints. We keep a threshold of minimum 10 matched keypoints to consider an image pair to be manipulated.

\noindent \textbf{DenseField:} We evaluated DenseField \cite{cozzolino2015efficient} on IDD task with three reported transforms - zernike moments (ZM), polar cosine transform (PCT) and fourier-mellin transform (FMT). ZM and PCT are evaluated with polar sampling grid. Feature length for ZM, PCT and FMT are 12, 10 and 25. Since DenseField is a copy-move detection algorithm, it expects a single image input. For evaluation on EDD task, we concatenated image pairs along the column axis to form a single input and used the best reported transform (ZM).

\noindent \textbf{DMVN:} 
The model is finetuned on synthetic data using adam optimizer with a learning rate of 1e-5, batch size 16 and binary crossentropy loss. The model has two outputs: 1) binary mask prediction and 2) image level forgery classification. We found fine-tuning to be unstable for joint training of both outputs. We set image classification loss weight to zero, tuning only the pixel loss. For image level classification we used the protocol similar to BusterNet \cite{Wu_2018_ECCV}. Postprocessing by removing stray pixels with less than 10\% of image area improved image classification performance.

\noindent \textbf{BusterNet:}
We finetune BusterNet \cite{Wu_2018_ECCV} on synthetic data using adam optimizer with a learning rate of 1e-5, batch size of 32 and categorical-crossentropy loss. BusterNet predicts a 3-channel mask to identify source, target and pristine pixels. Since we do not need to discriminate between source and target pixels, we consider both classes as manipulated.

\noindent \textbf{Block Feature Matching:} Discrete cosine transform (DCT), discrete wavelet transform (DWT) and Zernike features are matched with a block size of 16 pixels and minimum euclidean distance of 50 pixels between two matched blocks using the CMFD algorithm reported in \cite{christlein2012evaluation}.

\noindent \textbf{ManTraNet:}
We finetuned the model using adam optimizer with learning rate of 1e-3, batch size of 32 with gradient accumulation and binary-crossentropy loss. Since, cuts and transitions have thin pixel slices which can be distorted by resizing, we use images with original dimension.

\noindent \textbf{Baseline CNN:} We trained the CNN using adam optimizer with learning rate of 1e-3, mean squared error loss and batch size of 10.

\section{Alternate Metric: $F_{1}$ score}

Table \ref{duplication_detection} and \ref{copymove_detection} report $F_{1}$ scores for EDD and IDD tasks respectively. The experiments are identical to those reported in the main document, but with $F_{1}$ scores.
\section{Sample Predictions}

\begin{figure*}[!t]
    \centering
    \includegraphics[width=\linewidth]{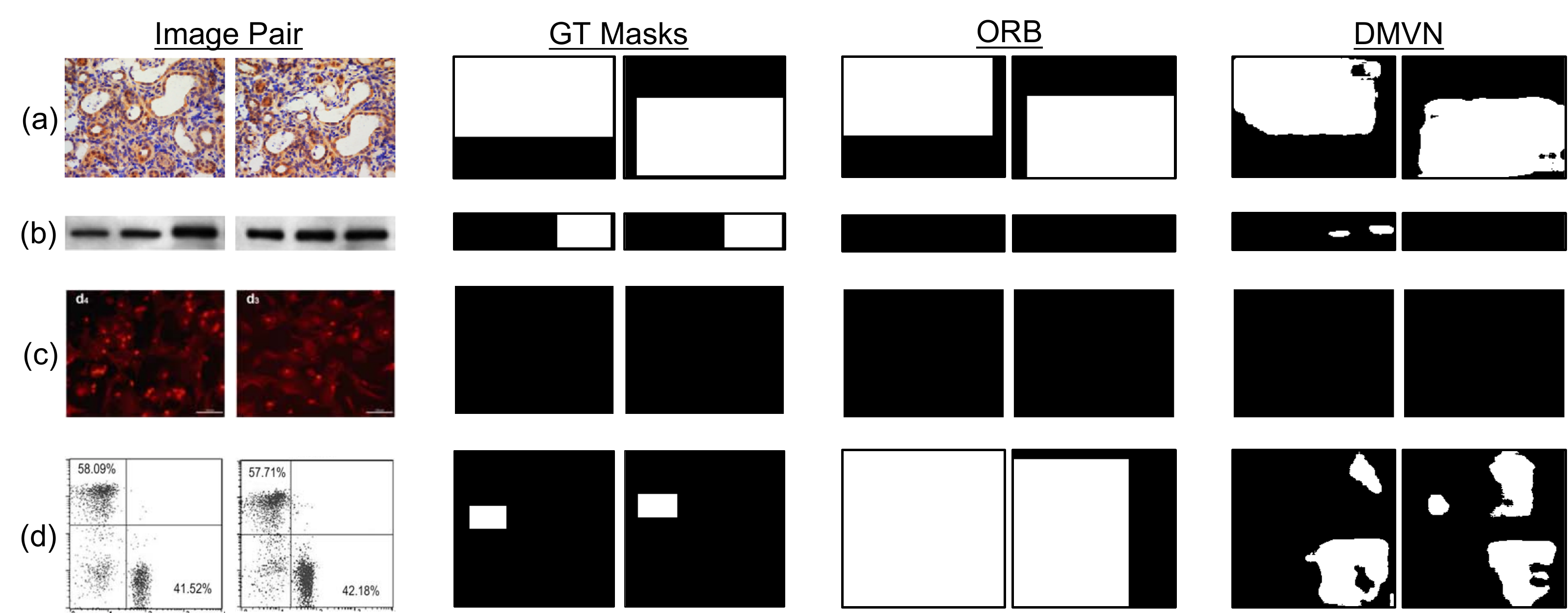}
    \caption{Rows of image pairs and corresponding predicted masks. The text in sample (d) misleads the prediction from both models.}
    \label{edd_pred}
\end{figure*}

\begin{figure*}[!t]
    \centering
    \includegraphics[width=\linewidth]{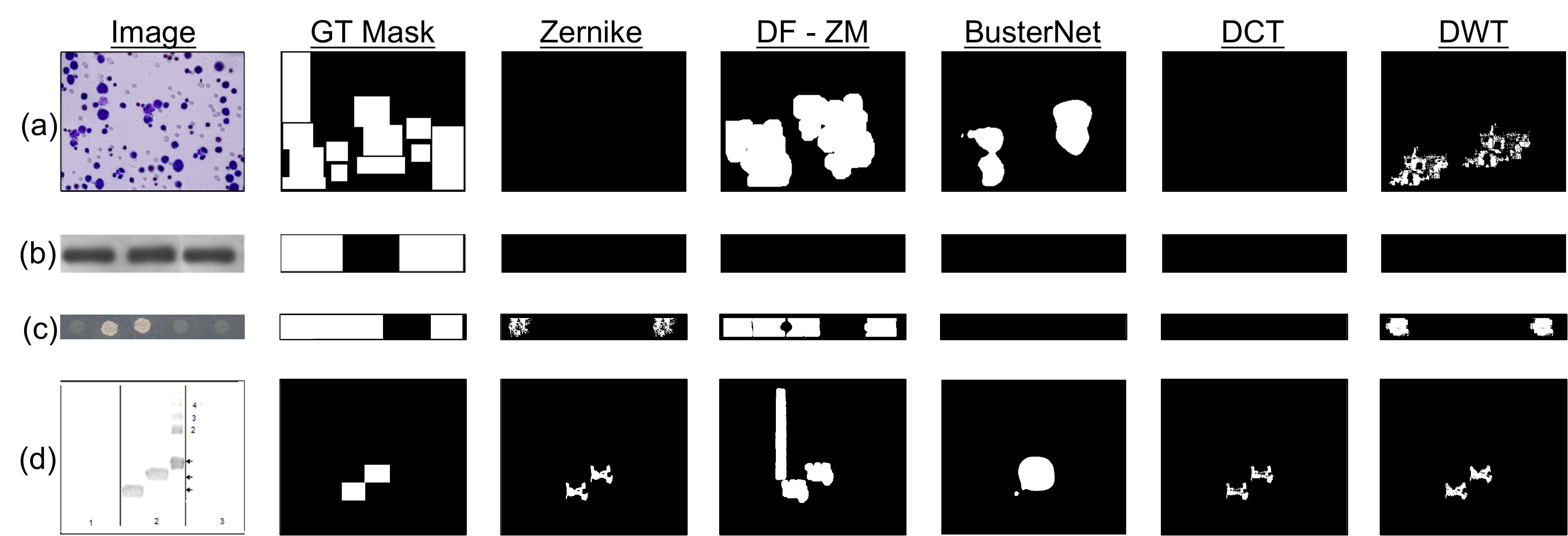}
    \caption{Rows of images and forgery detection predictions. There is significant variation in prediction across models. Rotated predictions in sample (a) are not identified by any model.}
    \label{idd_pred}
\end{figure*}

\begin{figure*}[!t]
    \centering
    \includegraphics[width=\linewidth]{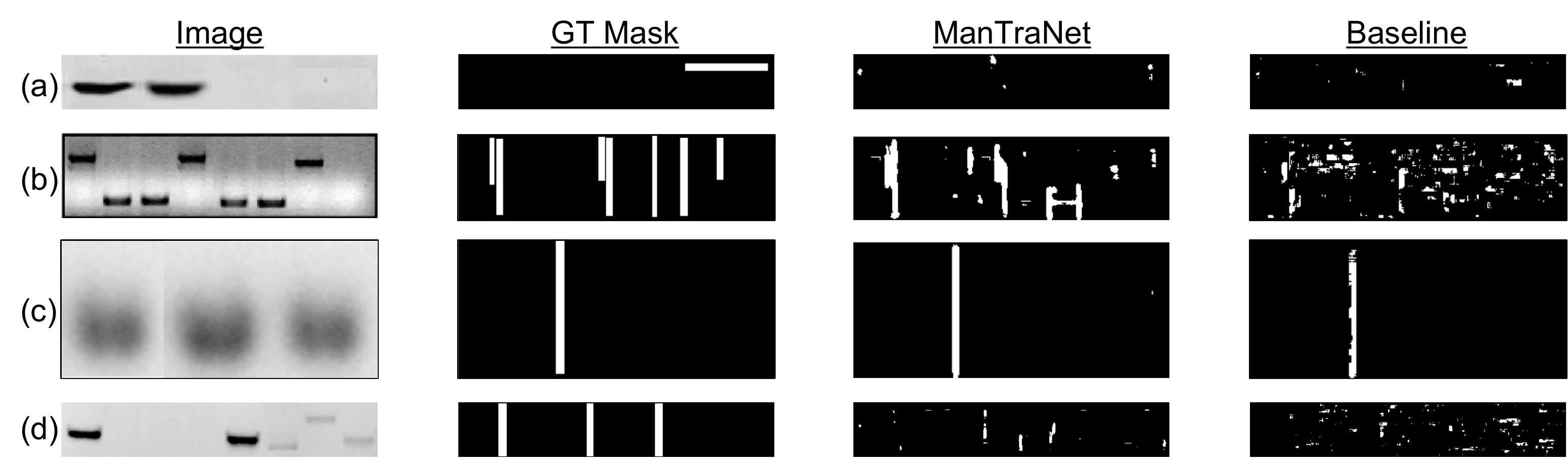}
    \caption{Predictions from ManTraNet and baseline CNN. It is evident that current forensic models are not suitable for the CSTD task.}
    \label{pcd_pred}
\end{figure*}

Prediction samples for EDD, IDD and CSTD tasks respectively in Figures \ref{edd_pred},\ref{idd_pred} and \ref{pcd_pred}. For EDD we show predictions from ORB \cite{rublee2011orb} and DMVN \cite{wu2017deep}. Samples for IDD include DCT \cite{fridrich2003detection}, DenseField \cite{cozzolino2015efficient}, DWT \cite{dwt}, Zernike \cite{ryu2010detection} and BusterNet \cite{Wu_2018_ECCV} baselines. Similarly, CSTD predictions are from ManTraNet \cite{Wu_2019_CVPR} and our cnn baseline. 

\section{Ethical Considerations}

We have used documents from PLOS to curate BioFors, since it is open access and can be used for further research including modification and distribution. However, the purpose of BioFors is to foster the development of algorithms to flag potential manipulations in scientific images. BioFors is explicitly not intended to malign or allege scientific misconduct against authors whose documents have been used. To this end, there are two precautions (1) We have anonymized images by withholding information about the source publications. Since scientific images have abstract patterns, matching documents from the web with BioFors images is a significant hindrance to the identification of source documents. (2) Use of pristine documents and documents with extenuating circumstances such as citation for duplication and justification. As a result, inclusion of a document in BioFors does not assure scientific misconduct.

\end{document}